\documentclass{article}

\usepackage[final]{neurips_2024}

\usepackage[utf8]{inputenc} 
\usepackage[T1]{fontenc}    
\usepackage{hyperref}       
\usepackage{url}            
\usepackage{booktabs}       
\usepackage{amsfonts}       
\usepackage{amsmath}
\usepackage{amssymb}
\usepackage{amsthm}
\usepackage{nicefrac}       
\usepackage{microtype}      
\usepackage{xcolor}         
\usepackage{graphicx}       
\usepackage{enumitem}
\usepackage{subfiles}       
\usepackage{framed}

\newtheorem{lemma}{Lemma}
\newtheorem{theorem}{Theorem}

\title{Catastrophic Goodhart: regularizing RLHF with KL divergence does not mitigate heavy-tailed reward misspecification}

\author{%
  Thomas Kwa \\
  Independent / FAR Labs \\
  \texttt{kwathomas0@gmail.com} \\
  \And
  Drake Thomas\thanks{Work was done while not at Anthropic} \\
  Anthropic \\
  \texttt{drake@anthropic.com} \\
  \And
  Adrià Garriga-Alonso \\
  FAR AI \\
  \texttt{adria@far.ai} \\
}

\begin{document}

\maketitle

 \begin{abstract}
  When applying reinforcement learning from human feedback (RLHF), the reward is learned from data and, therefore, always has some error. It is common to mitigate this by regularizing the policy with KL divergence from a base model, with the hope that balancing reward with regularization will achieve desirable outcomes despite this reward misspecification. We show that when the reward function has light-tailed error, optimal policies under less restrictive KL penalties achieve arbitrarily high utility. However, if error is heavy-tailed, some policies obtain arbitrarily high reward despite achieving no more utility than the base model--a phenomenon we call catastrophic Goodhart. We adapt a discrete optimization method to measure the tails of reward models, finding that they are consistent with light-tailed error. However, the pervasiveness of heavy-tailed distributions in many real-world applications indicates that future sources of RL reward could have heavy-tailed error, increasing the likelihood of reward hacking even with KL regularization.
\end{abstract}

\section{Introduction}

Kullback-Leibler (KL) divergence constraints in reinforcement learning (RL) are employed to stay in regimes where the objective is accurate enough.
Some on-policy \citep{trpo,ppo} and many off-policy \citep{abdolmaleki18mpo,jaques2019way} policy gradient algorithms employ KL constraints or penalties during optimization to prevent the policy from deviating too much from the data collection distribution. This ensures that estimates of each action's advantage are reliable enough to update the policy in a helpful way.

Reinforcement learning from human feedback \citep[RLHF]{rlhf_christiano,ziegler2020finetuning} is a very popular method to induce desirable behavior in language models. RLHF starts with a base pre-trained model, then learns a reward function from human annotator data. Next, it trains an RL policy to maximize this reward, while penalizing high KL divergence from the policy to the base model. RLHF uses an on-policy algorithm and has accurate advantages, but the \emph{reward function} is always somewhat misspecified compared to desired behavior, due to insufficient data, human biases, and other factors.

The main purpose of the KL penalty in RLHF is to limit the consequence of reward modeling errors by keeping the policy within a distribution similar to that on which it was trained. Ideally, in the low-KL regime the reward model's errors are small enough that it provides correct updates to the base model. \Citet{gao2023scaling} empirically supports this view: if the KL divergence in RLHF is allowed to grow too much, with a misspecified reward, the model's performance on the true utility starts to decrease.

We ask: can we obtain good outcomes from misspecified reward in RLHF by controlling the KL divergence? That is, if there is some error between the true reward $V$ and the proxy reward $U$, can the KL help us to still optimize $V$? Using mathematical proof, we answer the question in the negative for heavy-tailed errors: there exist policies which have infinite proxy reward $U$, but whose KL with the base model vanishes (these have undetermined $V$).  We term this phenomenon ``catastrophic Goodhart'', after Goodhart's law.

If the misspecification errors are independent and light-tailed, the KL divergence \emph{does} suffice to guarantee good outcomes. There may also be 
 guarantees under weaker assumptions, but assumptions that intuitively seem sufficient are often not (see Section~\ref{sec:discussion}).

Possibly, other regularization schemes would guarantee good outcomes for heavy-tailed errors, but this is not just a problem of KL. We show that optimizing by conditioning on large reward $U$ has similar outcomes in light- and heavy-tailed regimes.

Empirically, open-source language reward models seem to be light-tailed, which does not imply light-tailed errors but suggests it (Section~\ref{sec:experiments}). However, the errors are likely not independent and, given the prevalence of heavy-tailed distributions in the real world, error in future reward models may also be heavy-tailed. In any case, the present success of RLHF with misspecified rewards cannot be explained solely by the KL regularization in its objective.






\section{Background}

\subsection{KL divergence and KL regularization}

Recall that KL divergence between two distributions P and Q is defined as \[
D_{\mathrm{KL}}(P \| Q)=\sum_{x \in \mathcal{X}} P(x) \log \left(\frac{P(x)}{Q(x)}\right).
\]

If we have two policies $\pi, \pi_0$, we define $D_{KL}(\pi \| \pi_0)$ as the KL divergence between the distributions of actions taken on the states in trajectories reached by $\pi$. That is, if $Tr(\pi)$ is the distribution of trajectories taken by $\pi$, we penalize
\(D_{KL}(\pi \| \pi_0) \triangleq \mathbb E_{s \in T, T\sim Tr(\pi)}[D_{KL}(\pi(s) \| \pi_0(s))]\).

In RLHF, it is common to use the regularization term $\beta D_{KL}\left(\pi \| \pi_0 \right)$ to prevent the learned policy from deviating too much from the base policy, which can prevent unstable behavior or overfitting to the reward model. If our reward model gives reward $U$, then the optimal policy for RLHF with a KL penalty is
\[
\arg \max_{\pi} \mathbb{E} [U(\pi)]-\beta D_{KL}\left(\pi \| \pi_0 \right).
\]

Often the regularization parameter $\beta$ is dynamically adjusted to keep the $D_{KL}$ near some target value \citep{ziegler2020finetuning}.

\subsection{Heavy-tailed distributions}

A distribution $P$ over $\mathbb R$ with cumulative distribution function (CDF) $F_P$ is heavy-tailed if its tail function $\bar F_P(x) \triangleq 1 - F_P(x)$ satisfies 
\[
\lim _{x \rightarrow \infty} e^{t x} \bar{F}(x)=\infty \qquad \text{for all} t > 0.
\] 
Heavy-tailed distributions are well-known in statistics to have a higher probability of producing a single extreme value. For example, if the sum of two independent variables from heavy-tailed distributions is large, it is most likely due to one extreme sample rather than two equally large samples. \citep{wierman2013catastrophes}

\subsection{Reward misspecification and Goodhart's Law\label{sec:goodhart}}


Reward misspecification has caused low-utility outcomes in practice; for example, in \citep{ClarkAmodei2016}, an RL agent trained to play a racing videogame according to a misspecified reward function achieves a high score while failing to complete the course.

\citet{gao2023scaling} introduce the concept of ``overoptimization'': optimizing for a proxy objective decreases performance according to the true objective. This raises the question: in general, when RLHF reward is misspecified, when does the optimal policy produce high utility?

By applying the proxy reward and true reward functions to a distribution over text (generated by an LLM), we get two scalar random variables, which we call $U$ for proxy reward and $V$ for true reward / utility. Then we can define the error in the proxy reward as $X \triangleq U - V$, so that $U = X + V$. Framed this way, optimization for a proxy reward $U$ is a mix of desirable optimization for $V$ and undesirable optimization for $X$. The joint distribution of $V$ and $X$ determines the limiting value of $V$ as we apply more optimization. When we say that reward misspecification can have negative effects, we mean that too much variance in $X$ can "redirect" the optimization pressure from $V$ to $X$, and prevent utility gain from optimization.

Reward misspecification is also studied by \citep{lambert2024alignment}, \citep{laidlaw2024preventing}, and others. Laidlaw et al show that a KL penalty between action distributions can be ineffective, and propose instead regularizing state occupancy measure. Our results show an inherent weakness of KL divergence, including when applied to state occupancy measure.

We prove that in many cases, $V \to 0$ in the limit of optimization for some proxy $U$. We call this phenomenon ``catastrophic Goodhart'', after Goodhart's law: ``when a measure becomes a target, it ceases to be a good measure'' \citep{Strathern_1997}. In these cases, the end result of optimizing for a proxy of $V$ is no better than not optimizing at all. However, in other cases, $V \to \infty$ despite some reward misspecification; in these cases the reward misspecification is not severe enough to prevent good outcomes. 

\section{Theoretical results}

When applying KL regularization, the trained model is regularized towards some base policy $\pi_0$. One would hope that a KL penalty can produce good outcomes even in the case of reward misspecification; that is, if the reward $U$ is the sum of true utility $V$ and an error term $X$, we would hope that optimal policies under a KL penalty achieve high $V$ even if the magnitude of $X$ is large. We show that this is not always the case: Corollary~\ref{cor:heavy-tailed} of Theorems~\ref{thm1}, \ref{thm3}, and~\ref{thm2} establishes that when $X(\pi_0)$ is heavy-tailed, there are arbitrarily well-performing policies $\pi$ with $\mathbb E_{\pi}[V] \approx \mathbb E_{\pi_0}[V]$. However, Theorem~\ref{thm4} shows that when error is light-tailed and independent of $V$, the optimal policy under a KL penalty results in $V > 0$, and $V$ can be made arbitrarily large. Thus, the tails of the error distribution are crucial in determining how much utility will result from optimization towards an imperfect proxy.

Theorems~\ref{thm5} and \ref{thm6} (Section~\ref{sec:theoretical-conditioning}) show that the relationship of catastrophic Goodhart to heavy-tailed error is not just a quirk of KL divergence by using a different model of optimization based on conditioning on high reward values. Under this model (and given additional regularity conditions), it is also true that heavy-tailed error results in catastrophic Goodhart and light-tailed error plus independence results in arbitrarily large utility. All proofs are in the appendix.

\subsection{KL divergence on heavy- and light-tailed distributions}
\begin{theorem}
\label{thm1}
Given any heavy-tailed reference distribution
\(Q\) over \(\mathbb R\) with mean \(\mu_Q\), and any
\(M, \epsilon > 0\), there is a distribution \(P\) with mean \(\mu_P>M\)
and \(D_{KL}(P \| Q) < \epsilon\).
\end{theorem}

Outline of proof (see appendix for full proof): WLOG take $\mu_Q = 0$. If we set $P_t$ to upweight the probability mass of $Pr_{P_t}(X > t)$ to $c/t$ for some $c, t$, then the mean of $P_t$ will be approximately at least $c$. As $t \to \infty$, the KL divergence $D_{KL}(P_t \| Q)$ will shrink to zero.

Intuitively, in a heavy-tailed distribution, events with extremely high $x$ are not very rare, so you don't pay much of a KL penalty to upweight them so they happen about $1/x$ of the time. 

\begin{theorem}
    \label{thm2} However, if the distribution $Q$ is light-tailed and \(d = D_{KL}(P \| Q)\) is bounded, then
    \(\mathbb \mu_P \) is bounded, and \(\mu_P - \mu_Q \to 0\) as
    \(d \to 0\).
    \end{theorem}

\hypertarget{theorem-about-RLHF-with-KL-penalty}{%
\subsection{RLHF with KL penalty under heavy-tailed return distribution}}

We now adapt our result to the case where the policy is a language model and we are training it using RLHF. We are now applying KL divergence over the policies rather than the return distributions. We first formally define the properties of RLHF on language models that cause the result to hold: namely, when when considered as a Markov decision process (MDP), environmental transitions are deterministic and return depends only on the final state reached.

\textbf{\emph{Definition:}} A deterministic-transition MDP with
Markovian returns (DMRMDP) is an MDP \((\mathcal S, \mathcal A, P, R)\)
such that:

\begin{itemize}
\item
  The transition function
  \(P: \mathcal{S} \times \mathcal{A} \to \mathcal{S}\) is
  deterministic, i.e., for each state \(s \in \mathcal{S}\) and action
  \(a \in \mathcal{A}\), there exists a unique state
  \(s' \in \mathcal{S}\) such that \(P(s' | s, a) = 1\). \\ \textbf{In RLHF:} the transition is appending the generated token $a$ to the context $s$.
\item
  There is a set of sink states \(E \subseteq \mathcal S\) that
  terminate every trajectory, which is disjoint from the set of start
  states. \\ \textbf{In RLHF:} The sink states are sequences ending in \texttt{<EOS>} or above a certain length.
\item
  Returns are Markovian; that is, for any two trajectories
  \(\tau=(s_1, a_1, \dots, s_n), \tau'=(s_1', a_1', \dots, s_n'),\) if
  \(s_{n}= s'_{n}\), then \(\tau\) and \(\tau'\) have identical return
  distributions. Equivalently, for the trajectory random variable
  \(T=(S_1, A_1, \dots)\) distributed according to any policy, with
  return \(G\), \(G \bot\!\!\!\!\!\ \bot (S_{<i}, A_{<i}) \ |\ S_i\) for
  any \(i \ge 1\). \\ \textbf{In RLHF:} the return only depends on the full generated string, which is the final state.
\end{itemize}

The language model stochastically outputs the next token $a$ given $s$, and corresponds to the policy.
A DMRMDP is therefore a good model of RLHF.

\begin{theorem}
\label{thm3}
Let \(W = (\mathcal S, \mathcal A, P, R)\) be
a deterministic-transition MDP with Markovian returns. Given \(W\) we
define the function that takes policies to trajectories
\(Tr: (S \to \Delta A) \to \Delta(S \times A)^*\), and the average
return function \(g: (S \times A)^* \to \mathbb R\), which induces a
function \(G: \Delta(S \times A)^* \to \Delta \mathbb R\). Let
\(\pi_0: \mathcal S \to \Delta \mathcal A\) be some base policy. If
\(G \circ Tr(\pi_0)\) is heavy-tailed with finite mean \(\mu_Q\), then
for any \(M, \epsilon > 0\), there is a policy \(\pi\) with mean return
\(\mathbb E[U | U \sim G \circ Tr(\pi)] > M\) and
\(\mathbb E_{s \in T, T\sim Tr(\pi)}[D_{KL}(\pi(s) \| \pi_0(s))] < \epsilon\).
\end{theorem}

\newtheorem{corr}{Corollary}
\begin{corr}
\label{cor:heavy-tailed}
    Theorems~\ref{thm2} and \ref{thm3} imply that when utility is light-tailed, reward modeling errors make the proxy reward heavy-tailed, and a policy $\pi$ is regularized severely enough to have KL divergence values approaching zero, the reward $\mathbb E[U(\pi)]$ can go to infinity while utility $\mathbb E[V(\pi)]$ approaches a value no higher than the base policy.
\end{corr}

\hypertarget{light-tails-independence-imply-mathbb-ev-to-infty}{%
\subsection{\texorpdfstring{Light-tailed + independence imply
\(\mathbb E[V] \to \infty\)}{Light-tailed + independence imply \textbackslash mathbb EV \textbackslash to \textbackslash infty}}\label{light-tails-independence-imply-mathbb-ev-to-infty}}
 
\begin{theorem}
\label{thm4} If \(U=X+V\) with \(X\) and \(V\) both
light-tailed and \(V\) unbounded, and the distribution of U is continuous, and
\(\pi^*(\beta) \triangleq \arg \max_\pi \mathbb E[U(\pi)] - \beta D_{KL}(\pi, \pi_0)\),
then \(\lim_{\beta \to 0^+} \mathbb E[V(\pi^*(\beta))] = \infty\).
\end{theorem}

\subsection{Conditioning as alternate model of optimization\label{sec:theoretical-conditioning}}

Although we think a KL divergence penalty or cap is the most realistic setting for RLHF, it is not the only model of optimization where heavy-tailedness of the error determines whether catastrophic Goodhart occurs. Consider another model of optimization where $U = X+V$ as before, but we simply condition on $U$ being higher than some threshold $t$.\footnote{This could model a satisficing agent that takes random acceptable actions.} Then we are interested in the quantity $\lim_{t \to \infty} \mathbb E[V | X + V \ge t]$. If we slightly strengthen the heavy-tailedness and light-tailedness assumptions, heavy-tailed error results in catastrophic Goodhart, while light-tailed error results in arbitrarily high expected utility.

\subsubsection{Conditioning with heavy-tailed error produces catastrophic Goodhart}
\begin{theorem}
    \label{thm5} Let $X$ and $V$ be two independent random variables with CDFs $F_X$ and $F_V$ and tail functions $\bar F_V \triangleq 1 - F_V$, $\bar F_X \triangleq 1 - F_X$ such that
    \begin{itemize}
        \item $V$ has a finite mean.
        \item $X$ is subexponential; that is, $\lim_{x\to\infty}\frac{\text{Pr}(X_1+X_2>x)}{\text{Pr}(X>x)} = 2$ if $X_1, X_2$ are two independent samples from $X$. This is a slightly stronger property than being heavy-tailed.
        \item The tail of $V$ is sufficiently lighter than the tail of $X$ that \(\lim_{t\to\infty}\frac{t^p\bar F_V(t)}{\bar F_X(t)} = 0\) for some \(p > 1\).
    \end{itemize}
    Then $\lim_{t \to \infty} \mathbb E[V | X + V \ge t] =\mathbb{E}[V]$; that is, catastrophic Goodhart occurs in the limit of optimization for $U=X+V$.
\end{theorem}

The proof is included in the appendix. It requires expressing the conditional expectation in question as $\frac{\int_{-\infty}^\infty vf_V(v)\text{Pr}(X>t-v)} {\int_{-\infty}^\infty f_V(v)\text{Pr}(X>t-v)}$, then partitioning the interval $(-\infty, \infty)$ into four regions and bounding the integrand in the numerator above by a different quantity in each region. 

\subsubsection{Conditioning with light-tailed error produces arbitrarily high utility}
\begin{theorem}
    \label{thm6}
    Let $X, V$ be independent random variables such that $\lim_{t\to\infty}\frac{\bar{F}_X(t+1)}{\bar{F}_X(t)}=0$. (This implies that X has tails that are dominated by $e^{-cx}$ for any c, though it's a slightly stronger claim because it requires that X not have large jumps in the decay of its tails.)
    Then for any V with a finite mean which has no upper bound, $\lim_{t\to\infty}\mathbb{E}[V|X+V > t] = \infty$. 
\end{theorem}

Theorem 6 generalizes a consequence of the "Regressional Goodhart Identity" in \citep{gao2023scaling}.

\section{Experiments}

Our theoretical results now raise the question of whether the error in reward models is heavy-tailed or light-tailed in practice. \footnote{Note that distributions over a finite set are bounded and cannot be heavy-tailed in a technical sense, and models with a finite context window have a finite input space. For our purposes, a distribution of reward or error is effectively heavy-tailed if it takes on sufficiently large values and is well-modeled by a heavy-tailed distribution on its support.} If we observe the reward distribution to be light-tailed, this is a strong indication that error is light-tailed. \footnote{It is possible for $U$ to be light-tailed while $X$ and $V$ are both heavy-tailed, but this is unusual and we do not expect it to happen in practice.}

To empirically test whether the reward is heavy-tailed, we consider two lines of evidence: examining the distributions directly through random sampling and temperature-1 sampling, and finding adversarial token sequences that get high rewards. We examine one small and one medium reward model that performed reasonably well on RewardBench \citep{lambert2023rewardbench}. The small model is an OpenAssistant model based on Pythia 1.4B, and the medium model is Starling 7B-alpha \citep{starling2023}\footnote{https://huggingface.co/berkeley-nest/Starling-RM-7B-alpha}.

For random sampling, we sample 30000 length-1024 sequences of uniformly random tokens and observe the distribution of rewards assigned by both Pythia 1.4B and Llama 7B-chat. We also use Llama 7B-chat to generate 16000 length-133 sequences at temperature 1 and observe the distribution of rewards assigned by Starling 7B-alpha.

Because sampling is inefficient at probing the extreme tail, we also find token sequences that optimize Starling 7B-alpha for reward. We considered Greedy Coordinate Gradient (GCG) from \citep{zou2023universal}, a method used to find adversarial suffixes that circumvent jailbreaking, but decided on a faster version of GCG called Accelerated Coordinate Gradient (ACG) from \citep{haizelabs2024acg}. See Table \ref{tableh} for ACG hyperparameters.

\begin{table}
    \centering
    \label{tableh}
    \begin{tabular}{|c|c|}
        \hline
        Parameter & Value \\
        \hline
        Context length & 133 \\
        \hline
        Iterations & 1000 \\
        \hline
        Candidates per seq. position (k) & 3 \\
        \hline
        Annealing starting value & 9 \\
        \hline
        Annealing ending value & 2 \\
        \hline
    \end{tabular} \\
    \caption{Hyperparameters for ACG}
\end{table}

Generating plots took about 5 GPU-hours on 1x Nvidia H100, and running ACG took a further 8 hours.

\subsection{Results\label{sec:experiments}}

\begin{figure}
    \centering
    \includegraphics[width=0.8\linewidth]{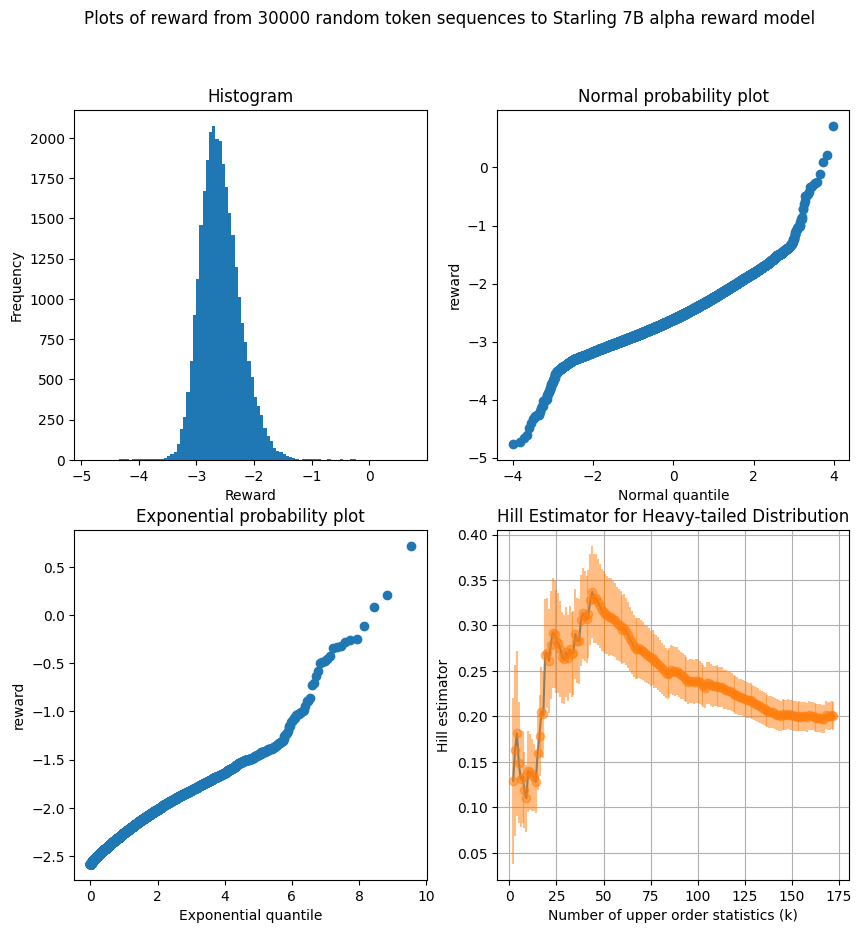}
    \caption{Plots of the distribution of reward from 30000 random length-1024 token sequences to Starling 7B-alpha. Clockwise from top left: The histogram shows a unimodal distribution with a slight right skew. The normal probability plot indicates the data are heavier-tailed than normal. The Hill estimator (error bars are standard error) appears to be 0.20 for higher values but fluctuates for lower values. The exponential probability plot of the right half of the distribution is consistent with either light or heavy tails (under heavy tails, the slope would go to infinity).}
    \label{fig:starling-random}
\end{figure}

\begin{figure}
    \centering
    \includegraphics[width=0.8\linewidth]{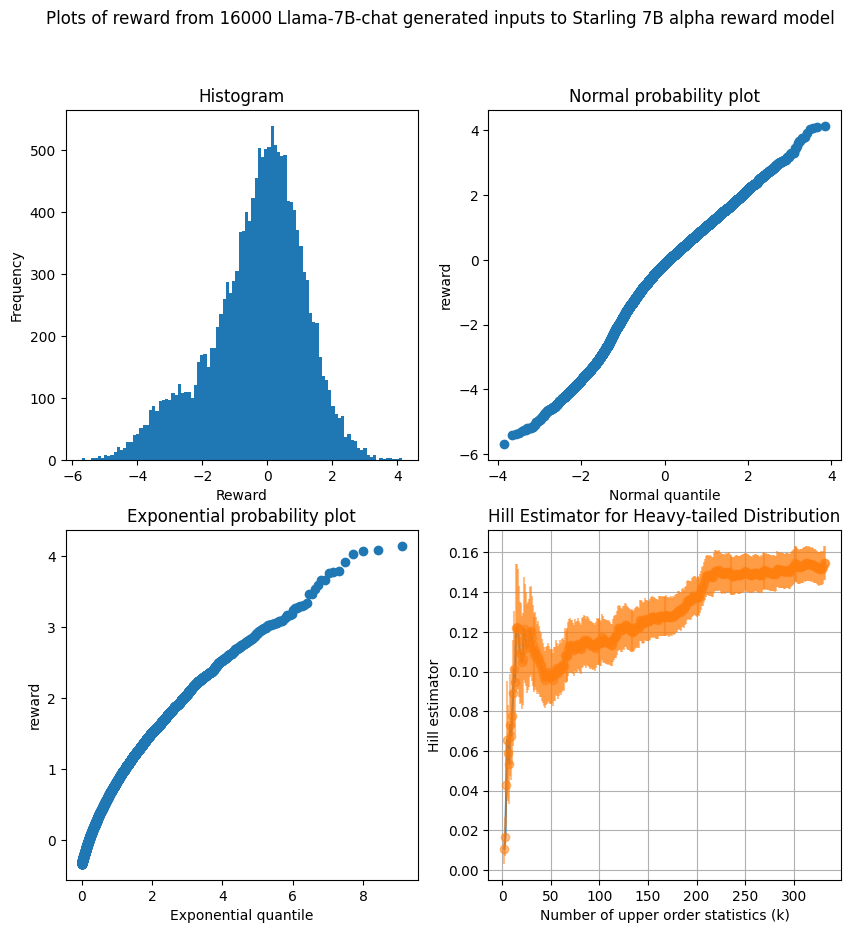}
    \caption{Plots of the reward distribution from 16000 token sequences generated by Llama 7B-chat of length $\le 133$, starting with five random tokens. Clockwise from top left: A histogram shows the reward distribution has a left skew. The normal probability plot suggests reward is approximately normal and thus light-tailed. The Hill estimator plot should stabilize if the distribution is heavy-tailed, but it does not; thus, there is no evidence the distribution is heavy-tailed. The exponential probability plot also indicates light tails, because the curve is bending downwards. }
    \label{fig:starling-llama}
\end{figure}

When sampling token sequences, both the Pythia model on random inputs (Figure \ref{fig:pythia-random}) and Starling 7B-alpha on Llama-generated inputs (Figure \ref{fig:starling-llama}) appear approximately normal and, therefore, light-tailed. Starling on random inputs (Figure \ref{fig:starling-random} is ambiguous, with the exponential Q-Q plot having an outlier that could indicate a heavy-tailed distribution, but the Hill estimator is consistent with a light-tailed distribution. Because Llama-7B-chat is a more reasonable base model than a completely random policy, we believe that Starling 7B-alpha is more likely to be light-tailed for the purposes of our theoretical results.

The ACG results need some interpretation. The KL divergence between two distributions $P$ and $Q$ if $P$ is the same as $Q$ a fraction $1-\alpha$ of the time, but is some value $x$ a fraction $\alpha$ of the time is given by \(D_{KL}(P \| Q) = \left[(1-\alpha)q(x) + \alpha\right] \log \left(\frac{(1-\alpha)q(x) + \alpha}{q(x)}\right) + 
(1-\alpha) \log (1-\alpha) (1 - q(x))\).

When $\alpha$ is small but much larger than $q(x)$, we approximate this to first order as $D_{KL}(P \| Q) \approx \alpha \log \left(\frac{\alpha}{q(x)}\right)$. In Theorems 1 and 3, we prove that when the error is sufficiently heavy-tailed, a policy that gets extremely large reward a small fraction of the time will achieve high expected reward with low KL divergence. This is not the case here because the rewards achieved through ACG were small and the log-probabilities extremely negative. For example, a policy that matches Llama 2-chat's base reward 99\% of the time and uses the highest-reward input generated by ACG $\alpha=$1\% of the time will have KL divergence from Llama 2-chat of $\alpha(\log(\alpha) - 1339.70) = 13.35$ nats, but reward only about $\alpha*(2.2377-0.3329) = 0.02571$ greater than the base model, far less than can be obtained with the same KL divergence by conditioning.

\section{Discussion and Limitations\label{sec:discussion}}

\subsection{How likely is catastrophic Goodhart?}

The low-KL policies that result in catastrophic Goodhart are not a unique optimal policy, just one family of high-performing policies. When optimizing $\mathbb{E} [U(\pi)]-\beta D_{K L}\left(\pi, \pi_0 \right)$, the outcome depends on RL training dynamics; it could be that $D_{KL} \to 0$ causing catastrophic Goodhart, but more likely both terms will go to infinity, potentially allowing $V \to \infty$. Catastrophic Goodhart can be prevented by using a light-tailed or bounded reward function.

Even so, catastrophic Goodhart is likely to occur in many scenarios where KL regularization is naively employed in an attempt to avoid Goodhart’s Law:

\begin{itemize}
    \item If we maximize $\sigma(\mathbb E[U]) + D_{KL}(Tr(\pi) \| Tr(\pi_0))$, where $\sigma$ is a bounded function (e.g. sigmoid), all near-optimal policies will have $V \approx 0$. Since we can only obtain so much reward from $\sigma(\mathbb{E}[U])$, it pays to make the KL (and thus V) go to zero.
    \item If we cap KL to a finite value (or dynamically adjust the KL penalty to target a finite KL, as done in \citet{ziegler2020finetuning}, then $\mathbb E[V]$ is also upper bounded by a finite value (see Theorem \ref{thm2}), and we think it is likely that $\mathbb E[V] \approx 0$. Consider a toy model where an AI can adjust three parameters: true quality $V$ of responses, frequency of reward hacking (producing actions with extremely high X), and severity of hacking (value of X on those actions). Adjusting the policy to increase $\mathbb E[U]$ without increasing KL increase the severity of hacking while decreasing either frequency of hacking or quality of responses. When $E[U]$ is already large, decreasing quality has much better returns than decreasing frequency. This is similar to Theorems \ref{thm5}, \ref{thm6} about hard-threshold optimization.
    \item Any way we maximize $\mathbb{E} [U(\pi)]-\beta D_{K L}\left(\pi, \pi_0 \right)$ results in very large values of $\mathbb E[U(\pi)]$, and there are a number of arguments that extreme optimization for an imperfect proxy can result in decreased utility due to tradeoffs between $X$ and $V$; e.g., the constrained resource scenario in \citep{zhuang2021consequences}.
\end{itemize}

\subsection{Independence assumptions}

Theorems 1-3 do not require any independence assumption, but Theorems \ref{thm4}, \ref{thm5}, and \ref{thm6} require that error $X$ and utility $V$ are independent, which seems to be violated in practice. Future work could weaken this assumption, although intuitively obvious ways to weaken it result in the statement being false. \footnote{Suppose that error $X$ is light-tailed conditional on any value of $V$, but our proxy is merely unbiased ($\mathbb E[X|V=v]=0$ for all $v$). Then the limit of $V$ under optimization for $X+V$ still depends on the relationship between $X$ and $V$. If they are independent, Theorem 6 says that \(\lim_{t \to\infty} \mathbb E[V | X + V \ge t] = \infty\). But if \(V \sim N(0, 1)\), and \(X | V \sim N(0, 4)\) when \(V \in [-1, 1]\), otherwise \(X=0\), then \(\lim_{t \to\infty} \mathbb E[V | X + V \ge t] = 0\).} 

\subsection{Stronger optimization methods}

We did not search the entire space of token sequences, so we cannot rule out that the reward is heavy-tailed enough to cause catastrophic Goodhart in some situations. While it is intractable to search the more than $10^{2000}$ possible token sequences, future work could get more evidence through more powerful optimization methods.

\subsection{Reparameterizing reward}

In some cases, a heavy-tailed reward can be reparameterized to make it light-tailed and avoid catastrophic Goodhart; however, in settings where the true reward is heavy-tailed, making reward artificially light-tailed or bounded can result in unintended behavior.

For example, a stock-trading agent should be rewarded by profit, but financial returns are known to be heavy-tailed. If we clip or otherwise transform rewards into a bounded interval, it will have no incentive to take into account huge gains or losses. Since RLHF rewards as implemented in Ziegler et al are unbounded, clipping or transforming rewards could itself cause reward misspecification.

In some cases, e.g. when the reward is not the true intended one, it is possible to reparameterize the reward without adverse effects. In the RL literature for Atari games, rewards are changes in score clipped to $[-1, 1]$ \citep{machado2018revisiting}.

\subsection{Relation to previous overoptimization work}

\citet{gao2023scaling} found that optimizing the reward of small reward models causes overoptimization: a decrease in utility with increasing optimization. However, we observed that reward models are light-tailed, and (Theorem \ref{thm4}) that independence combined with light-tailed error prevents overoptimization. We think this discrepancy is explained by dependence between error and utility. Policies optimized for high error may activate features in the proxy reward models that are undesirable according to the true utility function.\footnote{There are other explanations possible. Perhaps better optimization methods would find heavy-tailed reward in open reward models; or OpenAI's reward models have heavy-tailed error (and their results are straightforwardly explained by our Theorem~\ref{thm1}), while open reward models have light-tailed error.} 
More research is needed to understand why high-error completions have low utility and to design reward models that do not suffer from this problem; perhaps it is possible to construct reward models whose errors are in directions orthogonal to human preferences, so that the large-reward completions do not have lower utility.

\section{Conclusion}

We have argued that the purpose of the KL divergence regularization in RLHF is to mitigate reward misspecification. However, we have also proven that when errors in the reward function are heavy-tailed, it cannot serve this purpose: even with zero KL divergence, there are policies that achieve very high misspecified reward and no actual reward.

When errors are light-tailed and independent, the KL divergence can mitigate misspecification, but when they are dependent, this may not be possible. Thus, we must look to places other than the KL objective to explain the current success of RLHF and ensure its continued success in the future.

\begin{ack}
This work was supported by the Long-Term Future Fund (LTFF). We also thank the anonymous reviewers for their valuable feedback and constructive suggestions. 
\end{ack}

\section*{Impact Statement}

As this work aims to improve the safety of future ML systems by characterizing a possible failure mode of reward misspecification in RLHF, we hope the social impact is positive. We see no particular ethical issues to discuss.

\small{
\bibliographystyle{apalike} 
\bibliography{refs.bib} 
}


\appendix

\renewcommand\thefigure{\thesection.\arabic{figure}}    
\setcounter{figure}{0}

\renewcommand\thetable{\thesection.\arabic{table}}
\setcounter{table}{0}

\hypertarget{appendix}{%
\section{Proofs}\label{appendix}}

\subsection{Theorem 1}
\newtheorem*{theorem1}{Restatement of Theorem \ref{thm1}}
\begin{theorem1} Given any heavy-tailed reference distribution
\(Q\) over \(\mathbb R\) with mean \(\mu_Q\), and any
\(M, \epsilon > 0\), there is a distribution \(P\) with mean \(\mu_P>M\)
and \(D_{KL}(P \| Q) < \epsilon\).
\end{theorem1}

\begin{figure}
    \label{fig_thm1}
    \centering
    \includegraphics[width=0.3\linewidth]{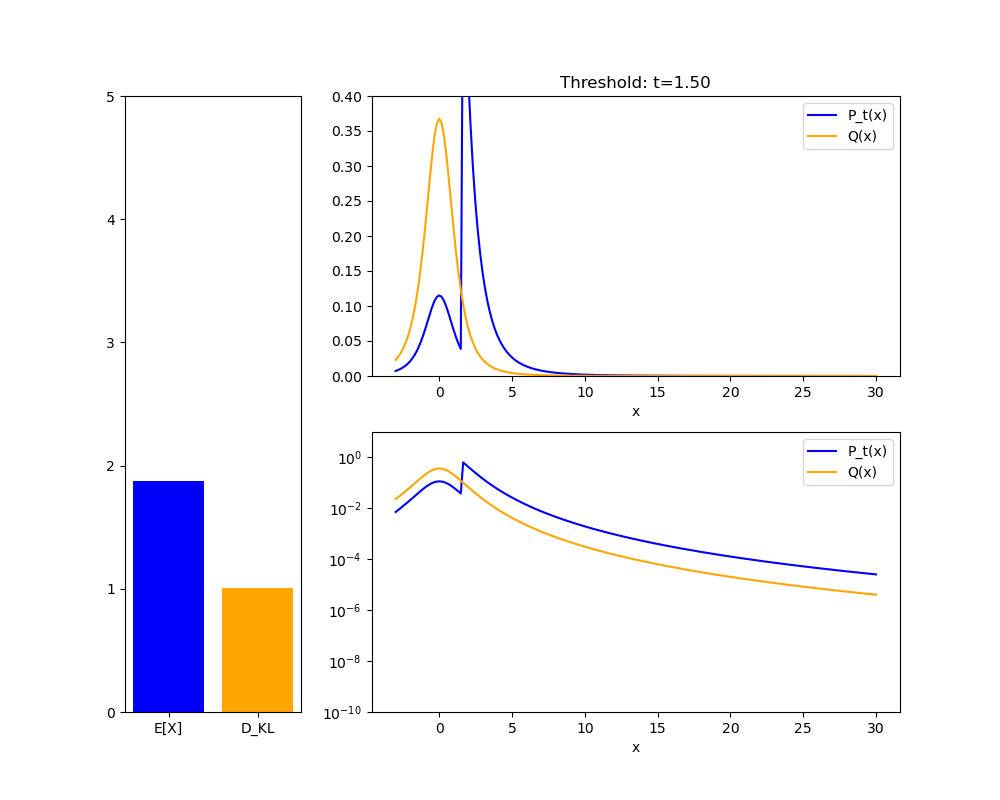}
    \includegraphics[width=0.3\linewidth]{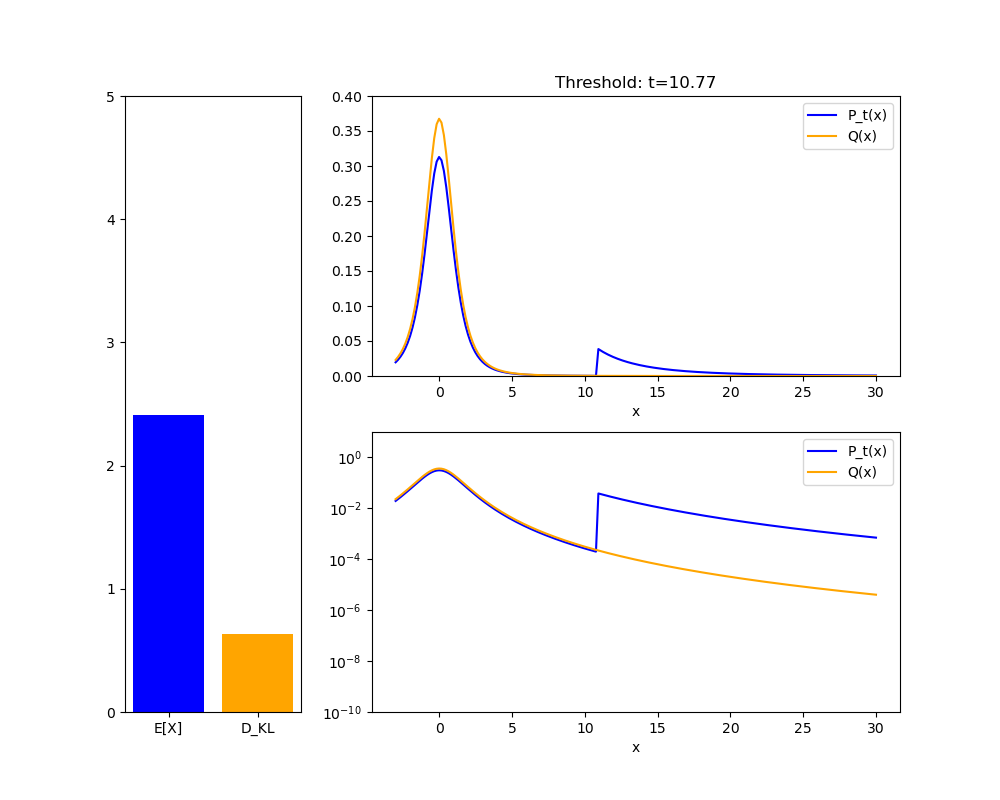}
    \includegraphics[width=0.3\linewidth]{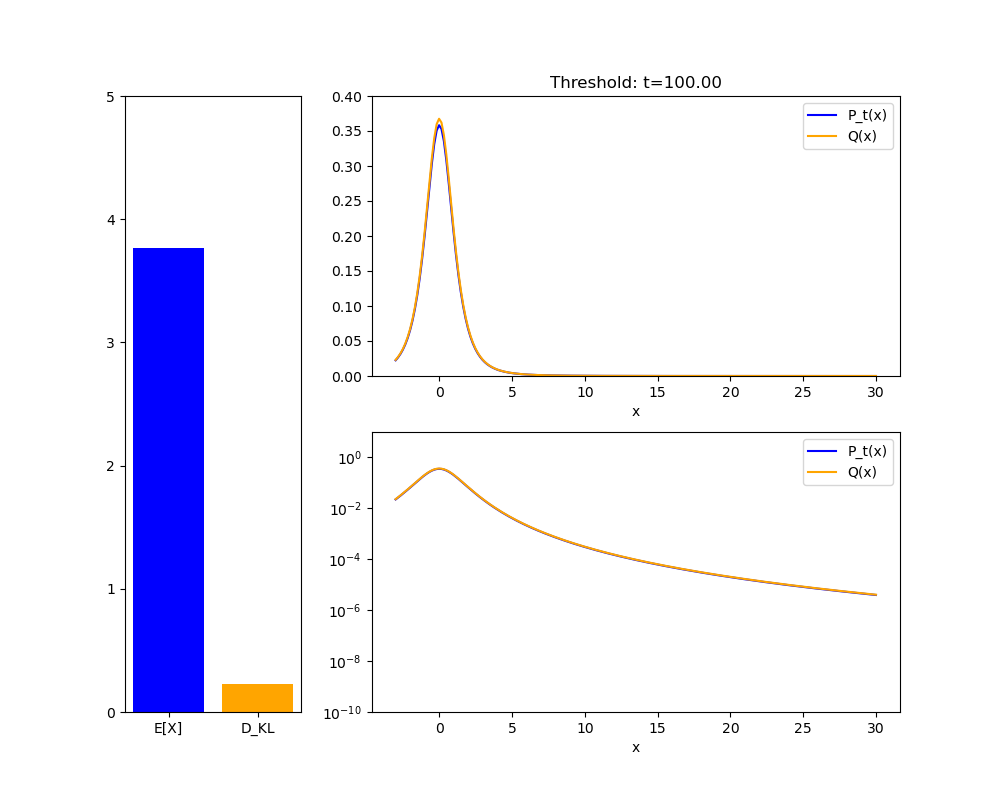}
    \caption{As $t \to \infty$, the mean of $X$ (blue bar) grows without bound while KL divergence $D_{KL}(P_t \,\|\, Q)$ (orange bar) goes to 0. The base distribution Q is a Student t-distribution with $df=3$. In this case, high values of X are upweighted to $1/t^{0.8}$; upweighting them to $1/t$ would cause $\mathbb E[X]$ to converge to ~$1$ while KL divergence goes to zero faster.}
\end{figure}

Intuitively, in a heavy-tailed distribution, events with extremely high $x$ are not very rare, so you don’t pay much of a KL penalty to upweight them so they happen about $1/x$ of the time. This is visually illustrated in Figure \ref{fig_thm1}.

\emph{Proof.} WLOG let \(\mu_Q = 0\). We construct a sequence of
distributions \(\{P_t\}\) such that
\(\lim_{t \to \infty} \mathbb E_{P_t}[X] \ge c\) for any constant \(c\),
and \(\lim_{t \to \infty} D_{KL}(P_t \| Q) = 0\). We define \(P_t\) for
any \(t > c\) thusly. Writing \(F_{P_t}(x)\) for the CDF
\(Pr_{X \sim P_t}(X \le x)\) and \(\bar F_{P_t}(x)\) for
\(1 - F_{P_t}(x)\), we let

$$ \bar F\_\{P\_t\}(x) =
\begin{cases} 1 - \frac{1 - c/t}{F_Q(t)}F_Q(x) & x \le t
\\ \frac{c/t}{\bar F_Q(t)}\bar F_Q(x) & x > t
\end{cases}
$$

Intuitively, we rescale the part of the distribution to the right of
\(t\) evenly to have total probability \(c/t\), which is less than 1
because \(t > c\).

We must check that \(\lim_{t \to \infty} \mathbb E_{P_t}[X] = c\). We
can write

\begin{align*} \mathbb E_{P_t}[X] = F_{P_t}(t) \mathbb E_{P_t}[X | X \le t] &+ \bar F_{P_t}(t) \mathbb E_{P_t}[X | X > t]    \\ = F_{P_t}(t) \mathbb E_Q[X | X \le t] &+ \bar F_{P_t}(t) \mathbb E_Q[X | X > t]    \\ = F_Q(t) \mathbb E_Q[X | X \le t] &+ \bar F_Q(t) \mathbb E_Q[X | X > t] + \\ (F_{P_t}(t) - F_Q(t))&E_Q[X | X \le t] + (\bar F_{P_t}(t) - \bar F_Q(t)) E_Q[X | X > t]     \\ = \mathbb E_Q[X] + (\bar F_{P_t}(t) - &\bar F_Q(t))(E_Q[X | X > t] - E_Q[X | X \le t])
\end{align*}

We know that \(\mathbb E_Q[X | X > t] > t\) because it is an integral of
values strictly greater than t. Because \(\mathbb E_Q[X] = 0\) is a
weighted average of \(\mathbb E_Q[X|X>t]\) and \(E_Q[X | X \le t]\), and
\(\mathbb E_Q[X|X>t] > 0\), we know \(E_Q[X | X \le t] < 0\). So
\(E_Q[X | X > t] - E_Q[X | X \le t] > t.\) We also know that for
sufficiently large \(t\), \((F_{P_t}(t) - F_Q(t)) > 0\). Intuitively,
starting from \(Q\), which has mean 0, \(P_t\) moves a probability mass
approaching \(\frac c t\) from mean \textless0 to mean \textgreater t.

Now we can say

\[
\lim_{t \to \infty} \mathbb E_{P_t}[X] > \lim_{t \to \infty}\left[ \mathbb E_Q[X] + (\bar F_{P_t}(t) - \bar F_Q(t))(t-0) \right]
\\ = \lim_{t \to \infty}\left( \frac c t - \bar F_Q(t) \right) t = \lim_{t \to \infty}c - t \bar F_Q(t)
\]

Because \(Q\) has a finite mean,
\(\lim_{t \to \infty} t \bar F_Q(t) = 0\), and so
\(\mathbb \lim_{t \to \infty} \mathbb E_{P_t}[X] \ge c\).

Now we check that \(\lim_{t \to \infty} D_{KL}(P_t \| Q) = 0\):

\begin{align*}D_{KL}({P_t} \| Q) &= \int_{\mathbb R} \log \frac{{P_t}(dx)}{Q(dx)} \,{P_t}(dx)\\&= \int_{x \le t} \log \frac{{P_t}(dx)}{Q(dx)} \,{P_t}(dx) + \int_{x > t} \log \frac{{P_t}(dx)}{Q(dx)} \,{P_t}(dx)\\ &= F_{P_t}(t) \log \frac{F_{P_t}(t)}{F_Q(t)} + \bar F_{P_t}(t) \log \frac{\bar F_{P_t}(t)}{\bar F_Q(t)} \text{\quad since both ratios are constant}\\ &= F_{P_t}(t) \log \frac{1 - c/t}{F_Q(t)} + \bar F_{P_t}(t) \log \frac{\bar F_{P_t}(t)}{\bar F_Q(t)}\end{align*}

Since both \(1-c/t\) and \(F_Q(t)\) go to \(1\) as \(t \to \infty\), the
left term goes to \(0\), and so

\begin{align*}\lim_{t \to \infty} D_{KL}(P_t \| Q)&\le 0 + \lim_{t \to \infty}\bar F_{P_t}(t) \log \frac{\bar F_{P_t}(t)
}{\bar F_Q(t)}
\\ & = \lim_{t \to \infty}\frac c t \log \frac c {t \bar F_Q(t)} \le \lim_{t \to \infty} \frac c t \log \frac 1 {\bar F_Q(t)} 
\\&= \lim_{t \to \infty}-\frac c t \log \bar F_Q(t) \text{\quad since t>c}
\end{align*}

\(Q\) is heavy-tailed, so by definition
\(\lim _{t \rightarrow \infty} e^{a t} \bar{F}_Q(t)=\infty \quad \text { for all } a>0\).
This implies that for every \(a > 0\) there is a sufficiently large
\(t_{c}\) so that for all \(t > t_c\), \(\bar F_Q(x) > e^{-at}\), which
means that \(\log \bar F_Q(t) > -a t\).

Therefore for every \(a > 0\),
\(\lim_{t \to \infty} D_{KL}(P_t \| Q) \le \lim_{t \to \infty} -c/t \log \bar F_Q(t) < \lim_{t \to \infty} -\frac {-act} t = ac\),
which since KL divergence is nonnegative means
that\(\lim_{t \to \infty} D_{KL}(P_t \| Q) = 0\) as desired.
\(\blacksquare\)

\subsection{Theorem 2}
\newtheorem*{theorem2}{Restatement of Theorem \ref{thm2}}
\begin{theorem2} If \(V\) is light-tailed, \(\mathbb E_Q[V]\)
is zero, and \(d = D_{KL}(P \| Q)\) is bounded, then
\(\mathbb E_P[V]\) is bounded, and \(\mathbb E_P[V] \to 0\) as
\(d \to 0\).
\end{theorem2}

\emph{Proof.} Using Lagrange multipliers, we find that when KL divergence is
minimized, we have
\(P(V) [\lambda_1 \log \frac{P(V)}{Q(V)} + \lambda_2 - X] = 0\) for some
constants \(\lambda_1, \lambda_2\), so

\begin{align}\log \frac{P(V)}{Q(V)} = \frac{V - \lambda_2}{\lambda_1}
\\ P(V) = Q(V)\exp\left(\frac{V - \lambda_2}{\lambda_1}\right) = Q(V)
\\ e^{V/\lambda\_1} e^{-\lambda\_2/\lambda\_1} = C Q(V)
e^{V/\lambda\_1}
\end{align}

That is, the new PDF is an
\href{https://en.wikipedia.org/wiki/Exponential_tilting}{exponential
tilting} of the old PDF. Now, what is \(\mathbb E_P[V]\)? It's just
\(\int_{-\infty}^{\infty} C V e^{V/\lambda_1} Q(X) \,dV\). If the
distribution of V is heavy-tailed distribution, this is \(\infty\); if
it is light-tailed, this is some finite value.

When \(d = 0\), \(P\) and \(Q\) are identical, and \(\mathbb E[V] = 0\).
So by a continuity argument, \(\mathbb E_P[V] \to 0\) as \(d \to 0\).
\(\blacksquare\)

\subsection{Theorem 3}
\newtheorem*{theorem3}{Restatement of Theorem \ref{thm3}}
\begin{theorem3}
Let \(W = (\mathcal S, \mathcal A, P, R)\) be
a deterministic-transition MDP with Markovian returns. Given \(W\), we
define the function that takes policies to trajectories
\(Tr: (S \to \Delta A) \to \Delta(S \times A)^*\), and the average
return function \(g: (S \times A)^* \to \mathbb R\) which induces a
function \(G: \Delta(S \times A)^* \to \Delta \mathbb R\). Let
\(\pi_0: \mathcal S \to \Delta \mathcal A\) be some base policy. If
\(G \circ Tr(\pi_0)\) is heavy-tailed with finite mean \(\mu_Q\), then
for any \(M, \epsilon > 0\), there is a policy \(\pi\) with mean return
\(\mathbb E[U | U \sim G \circ Tr(\pi)] > M\) and
\(\mathbb E_{s \in T, T\sim Tr(\pi)}[D_{KL}(\pi(s) \| \pi_0(s))] < \epsilon\).
\end{theorem3}

\emph{Proof:} We will exhibit a distribution of trajectories \(\rho\)
such that \(D_{KL}(\rho \| Tr(\pi_0)) < \epsilon\) and
\(\mathbb E[G(\rho)] > M\), and then construct a policy \(\pi\) with
\(Tr(\pi) = \rho\). Note that this proof applies for continuous action
spaces if trajectories are replaced with measurable sets, but this would
make it harder to read.

Let \(\rho_{\pi_0} = Tr(\pi_0)\). We have a heavy-tailed distribution of
return \(Q \triangleq G(\rho_{\pi_0})\) over \(\mathbb R\), so we can
apply Theorem~\ref{thm1}. But to define \(\rho\), we can construct \(P_t\) in the
proof of Theorem~\ref{thm1} in a particular way. For any \(t>c\), we need a
\(P_t\) that uniformly upweights values of mean return such that
\(\bar F_{P_t}(t) = c/t\). We can define \(\rho_t\) such that any
trajectory \(\tau\) is upweighted by a factor depending only on its mean
return:

\[
\rho_t(\tau) = \begin{cases} \frac{1 - c/t}{F_Q(t)} \rho_{\pi_0}(\tau) & g(\tau) \le t
\\ \frac{c/t}{\bar F_Q(t)}\rho_{\pi_0}(\tau) & g(\tau) > t
\end{cases}
\]

Then we can let \(P_t \triangleq G \circ \rho_t\) and the rest of the
proof of Theorem~\ref{thm1} applies. Therefore, applying the theorem, we can let
\(\rho = \rho_t\) for sufficiently large \(t\), and then
\(\mu_{G \circ \rho} > M\) and
\(D_{KL}(G \circ \rho, G \circ \rho_{\pi_0}) < \epsilon\). By the
chain rule for KL divergence,
\(D_{KL}(\rho, \rho_{\pi_0}) = D_{KL}(G\circ \rho, G \circ \rho_{\pi_0}) + \mathbb E_{\gamma \sim G\circ\rho}[D_{KL}(\rho(T) | G(T)=\gamma \ \|\ \rho_{\pi_0}(T) | G(T)=\gamma)]\).
Since we constructed \(\rho\) so that the probabilities of each \(\tau\)
conditional on its return being \(\gamma\) are equal, the second term is
zero, and we also have \(D_{KL}(\rho, \rho_{\pi_0}) < \epsilon\).

Finally, since the KL divergence between trajectory distributions is the
sum of KL divergence between policies at each action in the trajectory,
and each trajectory has at least one action,
\(\mathbb E_{s \in T, T\sim Tr(\pi)}[D_{KL}(\pi(s) \| \pi_0(s))] \le \mathbb E_{T\sim Tr(\pi)} \sum_{s \in T}[D_{KL}(\pi(s) \| \pi_0(s))] = D_{KL}(\rho \| \rho_{\pi_0}) < \epsilon\)
as desired.

To define \(\pi\) such that \(Tr(\pi) = \rho\), we let
\(\pi(s, a) = Pr(a_i = a | \tau = (..., s, a_i, ...) \sim \rho)\).

Then, the probability that any trajectory
\(\tau = (s_1, a_1, \dots, a_n)\) is sampled is:

\begin{align}
Tr(\pi)(\tau) &= \prod_{i=1}^n \pi(s_i, a_i) 
\\&= \prod_{i=1}^n Pr(a_i=a_i' | \tau' = (..., s, a_i', ...) \sim \rho)
\\&= \prod_{i=1}^n Pr(a_i = a_i' | \tau' = (s_1', a_1', ..., s, a_i', ...) \sim \rho, s_{<i} = s'_{<i}, a_{<i} = a'_{<i})
\\&= \rho(\tau)
\end{align}

In (2), returns are Markovian, so all trajectory prefixes ending in
state \(s\) have the same distribution of returns under any policy. In
the construction of \(\rho\), all trajectories with the same mean return
have equal measure. Therefore, conditioning on earlier states and
actions of \(\tau\) does not change the measure, so we can write (3). So
\(Tr(\pi)=\rho\) as desired. \(\blacksquare\)

\subsection{Theorem 4}

\newtheorem*{theorem4}{Restatement of Theorem \ref{thm4}}
\begin{theorem4}
    If \(U=X+V\) with \(X\) and \(V\) both
    light-tailed and \(V\) unbounded, and the distribution of U is continuous, and
    \(\pi^*(\beta) \triangleq \arg \max_\pi \mathbb E[U(\pi)] - \beta D_{KL}(\pi, \pi_0)\),
    then \(\lim_{\beta \to 0^+} \mathbb E[V(\pi^*(\beta))] = \infty\).
\end{theorem4}

\emph{Proof.} Fix some \(\beta\). Using Lagrange multipliers, we find
that for any event \(S\),
\(\Pr_\pi(S) = \Pr_{\pi_0}(S) e^{\lambda U(S)}\). Let \(c(\beta)\) be
the median value of \(U\) under the policy \(\pi^*(\beta)\); that is,
\(Pr(U > c(\beta) | U \sim G \circ Tr(\pi^*(\beta))) = \frac 1 2.\) This
exists because \(U\) has a continuous distribution. Then:

\begin{align*}E[V | \pi] &= \frac 1 2 E[V | \pi, U < c] + \frac 1 2 E[V | \pi, U \ge c]
\\ &\ge \frac 1 2 E[V | \pi, U < c] + \frac 1 2 E[V | \pi]
\\ \lim_{\beta \to 0^+} E[V | \pi] &\ge \lim_{\beta \to 0^+} \frac 1 2 E[V | \pi, U < c] + \lim_{\beta \to 0^+} \frac 1 2 E[V | \pi]
\end{align*}

The left term is \(c\), while the right term is \(\infty\), so the
overall limit is \(\infty\).

\subsection{Theorem 5}

\newtheorem*{theorem5}{Restatement of theorem {\ref{thm5}}}
\begin{theorem5}
    Let $X$ and $V$ be two independent random variables with CDFs $F_X$ and $F_V$ and tail functions $\bar F_V \triangleq 1 - F_V$, $\bar F_X \triangleq 1 - F_X$ such that
    \begin{itemize}
        \item $V$ has a finite mean.
        \item $X$ is subexponential; that is, $\lim_{x\to\infty}\frac{\text{Pr}(X_1+X_2>x)}{\text{Pr}(X>x)} = 2$ if $X_1, X_2$ are two independent samples from $X$. This is a slightly stronger property than being heavy-tailed.
        \item The tail of $V$ is sufficiently lighter than the tail of $X$ that \(\lim_{t\to\infty}\frac{t^p\bar F_V(t)}{\bar F_X(t)} = 0\) for some \(p > 1\).
    \end{itemize}
    Then $\lim_{t \to \infty} \mathbb E[V | X + V \ge t] =\mathbb{E}[V]$; that is, catastrophic Goodhart occurs in the limit of optimization for $U=X+V$.
\end{theorem5}

The proof requires expressing the conditional expectation in question as $\frac{\int_{-\infty}^\infty vf_V(v)\text{Pr}(X>t-v)} {\int_{-\infty}^\infty f_V(v)\text{Pr}(X>t-v)}$, then partitioning the interval $(-\infty, \infty)$ into four regions and bounding the integrand in the numerator above by a different quantity in each region.

In addition to the works cited in the main paper, we make reference to the textbook \citep{foss2013introduction} throughout the proof. Many similar results about random variables are present in the textbook.

\subsubsection{Proof sketch and intuitions}

 The conditional expectation \(\mathbb E[V | X + V > t]\) is given by \(\frac{\int_{-\infty}^\infty vf_V(v)\text{Pr}(X>t-v)} {\int_{-\infty}^\infty f_V(v)\text{Pr}(X>t-v)}\), \footnote{We'll generally omit \(dx\) and \(dv\) terms in the interests of compactness and conciseness; the implied differentials should be pretty clear.}
 and we divide the integral in the numerator into 4 regions, showing that each region's effect on the conditional expectation of V is similar to that of the corresponding region in the unconditional expectation \(\mathbb E[V]\).

The regions are defined in terms of a slow-growing function \(h(t):\mathbb R \to \mathbb R_{\ge 0}\) such that the fiddly bounds on different pieces of the proof work out. Roughly, we want it to go to infinity so that \(|V|\) is likely to be less than \(h(t)\) in the limit, but grow slowly enough that the shape of \(V\)'s distribution within the interval \([-h(t),h(t)]\) doesn't change much after conditioning.

In Table \ref{table1}, we abbreviate the condition \(X+V>t\) as \(c\).

\begin{table}
\label{table1}
    \centering
    \begin{tabular}{|c|c|p{70mm}|}
    \hline \\
         Region
        &Why its effect on \(\mathbb E[V|c]\) is small
        & Explanation
    \\ \hline \\
        \(r_1=(-\infty,-h(t)]\)
        & \(\mathbb P[V\in r_1 | c]\) is too low  & In this region, \(|V| > h(t)\) and \(X > t + h(t)\), both of which are unlikely.
    \\ \hline \\
        \(r_2=(-h(t),h(t))\) 
        &  \(\mathbb E[V|V \in r_2, c] \approx \mathbb E[V|V \in r_2]\)
        & The tail distribution of X is too flat to change the shape of \(V\) 's distribution within this region.
    \\ \hline \\
        \(r_3\!=\![h(t),t\!-\!h(t)]\)
        & \(\mathbb P\left[ V \in r_3\ |\ c\right]\) is low, and \(V<t\).
        & There are increasing returns to each bit of optimization for X, so it's unlikely that both X and V have moderate values. \footnote{The diagrams in the previous post TODO show visually that when \(X\) and \(V\) are both heavy-tailed and \(t\) is large, most of the probability mass has \(X \approx 0\), \(V \approx t\) or vice versa.}
    \\ \hline \\
    \(r_4=(t-h(t),\infty)\) &  \(\mathbb P[V\in r_4 \ |\ c]\) is too low
        & X is heavier-tailed than V, so the condition that \(V > t-h(t)\) is much less likely than \(X > t - h(t)\) in \(r_2\).
    \\ \hline
    \end{tabular}
    \caption{A summary of the proof strategy for Theorem 5.}
\end{table}

\begin{figure}
    \centering
    \includegraphics[width=0.5\linewidth]{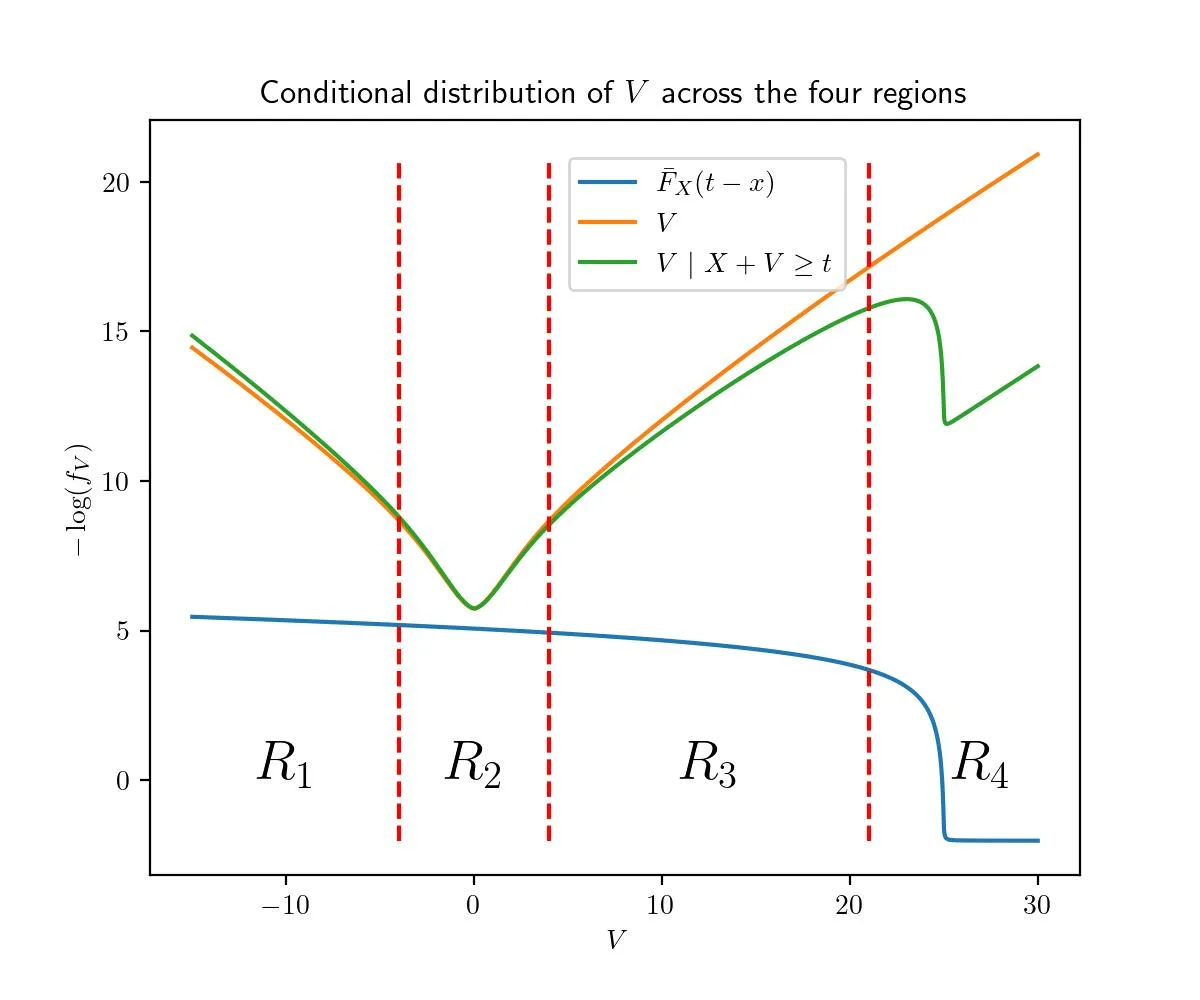}
    \caption{A diagram showing the region boundaries at \(-h(t)\), \(h(t)\), and \(t-h(t)\) in an example where \(t=25\) and \(h(t)=4\), along with a negative log plot of the relevant distribution:}
    \label{fig:theorem5-diagram}
\end{figure}

Note that up to a constant vertical shift of normalization, the green curve is the pointwise sum of the blue and orange curves.

\subsubsection{Definitions}

To be more precise, we're going to make the following definitions and assumptions:

Let \(f_V(v)\) be the PDF of \(V\) at the value \(v\). We assume for convenience that \(f_V\) exists, is integrable, etc, though we suspect that this isn't necessary, and that one could work through a similar proof just referring to the tails of \(V\). We won't make this assumption for \(X\).
Let \(F_X(x)=\text{Pr}(X\le x)\) and \(\bar{F}_X(x)=\text{Pr}(X>x)\), similarly for \(F_V\) and \(\bar{F}_V\). 
Assume that
\begin{itemize}
    \item \(V\) has a finite mean: \(\int_{-\infty}^\infty vf_V(v)\,dv\) converges absolutely.
    \item \(X\) is subexponential.
\end{itemize}

Formally, this means that \(\lim_{x\to\infty}\frac{\text{Pr}(X_1+X_2>x)}{\text{Pr}(X>x)} = 2\). 
This occurs roughly whenever \(X\) has tails that are heavier than \(e^{-cx}\) for any \(c\) and is reasonably well-behaved; counterexamples to the claim "long-tailed implies subexponential" exist, but they're nontrivial to exhibit.
Examples of subexponential distributions include log-normal distributions, anything that decays like a power law, the Pareto distribution,and distributions with tails asymptotic to \(e^{-x^a}\) for any \(0<a<1\).

We require for \(V\) that its tail function is substantially lighter than X's, namely that \(\lim_{t\to\infty}\frac{t^p\bar F_V(t)}{\bar F_X(t)} = 0\) for some \(p > 1\). (This implies that \(\bar F_V(t) = O(\bar F_X(t) / t)\).)

With these definitions and assumptions, we can move on to the proof. 

The unnormalized PDF of \(V\) conditioned on \(X+V\ge t\) is given by \(f_V(v)\bar{F}_X(t-v)\). Its expectation is given by \(\frac{\int_{-\infty}^\infty vf_V(v)\bar{F}_X(t-v)} {\int_{-\infty}^\infty f_V(v)\bar{F}_X(t-v)}\).

Meanwhile, the unconditional expectation of V is given by \(\int_{-\infty}^\infty vf_V(v)\).

We'd like to show that these two expectations are equal in the limit for large \(t\). To do this, we'll introduce \(Q(v)=\frac{\bar{F}_X(t-v)}{\bar{F}_X(t)}\). (More pedantically, this should really be \(Q_t(v)\), which we'll occasionally use where it's helpful to remember that this is a function of \(t\).)

For a given value of \(t\), \(Q(v)\) is just a scaled version of \(\bar{F}_X(t-v)\), so the conditional expectation of \(V\) is given by \(\frac{\int_{-\infty}^\infty vf_V(v)Q(v)} {\int_{-\infty}^\infty f_V(v)Q(v)}\). But because \(Q(0)=1\), the numerator and denominator of this fraction are (for small \(v\) ) close to the unconditional expectation and \(1\), respectively.

We'll aim to show that for all \(\epsilon>0,\) we have for sufficiently large \(t\) that \(\left|\int_{-\infty}^\infty vf_V(v)Q_t(v) - \int_{-\infty}^\infty vf_V(v)\right|<\epsilon\) and \(\int_{-\infty}^\infty f_V(v)Q_t(v) \in [1-\epsilon,1+\epsilon]\), which implies (exercise) that the two expectations have limiting difference zero. But first we need some lemmas.

\subsubsection{Lemmas}

\begin{lemma}
 There is \(h(t)\) depending on \(F_X\) such that:
\begin{enumerate}[label=(\alph*)]
    \item \(\lim_{x\to\infty} h(t)=\infty\)
    \item \(\lim_{t \to\infty} t - h(t) = \infty\)
    \item \(\lim_{t\to\infty}\frac{\bar F_X(t-h(t))}{\bar F_X(t)}=1\)
    \item \(\lim_{t \to \infty} \sup_{|v| \le h(t)} |Q(v,t)-1| = 0\).
\end{enumerate}
\end{lemma}

\begin{proof}
 Lemma 2.19 from \citep{foss2013introduction} implies that if \(X\) is long-tailed (which it is, because subexponential implies long-tailed), then there is \(h(t)\) such that condition (a) holds and \(\bar F_X\) is \(h\)-insensitive; by Proposition 2.20 we can take \(h\) such that \(h(t) \le t/2\) for sufficiently large \(t\), implying condition (b). Conditions (c) and (d) follow from being \(h\)-insensitive.
 \end{proof}

\begin{lemma}
 Suppose that \(F_X\) is whole-line subexponential and \(h\) is chosen as in Lemma 1. Also suppose that \(\bar F_V(t) = O(\bar F_X(t) / t)\). Then \(Pr[X+V>t,\ V>h(t),\ X > h(t)] = o(\bar F_X(t)/t).\)
 \end{lemma}
 
\begin{proof}

 This is a slight variation on lemma 3.8 from \citep{foss2013introduction}, and follows from the proof of Lemma 2.37. Lemma 2.37 states that

\begin{quote}
    \textbf{Lemma 2.37.} Let $h$ be any increasing function on $\mathbb{R}^{+}$such that $h(x) \rightarrow \infty$. Then, for any distributions $F_1, F_2, G_1$, and $G_2$ on $\mathbb{R}$,
$$
\limsup _{x \rightarrow \infty} \frac{\mathbb{P}\left\{\xi_1+\eta_1>x, \xi_1>h(x), \eta_1>h(x)\right\}}{\mathbb{P}\left\{\xi_2+\eta_2>x, \xi_2>h(x), \eta_2>h(x)\right\}} \leq \limsup _{x \rightarrow \infty} \frac{\overline{F_1}(x)}{\overline{F_2}(x)} \cdot \limsup _{x \rightarrow \infty} \frac{\overline{G_1}(x)}{\overline{G_2}(x)},
$$
where $\xi_1, \xi_2, \eta_1$, and $\eta_2$ are independent random variables with respective distributions $F_1, F_2, G_1$ and $G_2$.
\end{quote}

but it is actually proved that \begin{multline}\mathbb{P}\left\{\xi_1+\eta_1>x, \xi_1>h(x), \eta_1>h(x)\right\} \leq \\ \sup_{z > h(x)} \frac{\overline{F_1}(z)}{\overline{F_2}(z)} \cdot \sup_{z > h(x)} \frac{\overline{G_1}(z)}{\overline{G_2}(z)} \cdot {\mathbb{P}\left\{\xi_2+\eta_2>x, \xi_2>h(x), \eta_2>h(x)\right\}}.\end{multline}

If we let \(F_1 = F_V, F_2=G_1=G_2=F_X\), then we get  \begin{multline} \mathbb{P}\left\{X+V>t, X>h(t), V>h(t) \right\}  \\ \le \sup_{z > h(t)} \frac{\bar F_V(z)}{\bar F_X(z)} \sup_{z > h(t)} \frac{\bar F_X(z)}{\bar F_X(z)} {\mathbb{P}\left\{X+X'>t, X>h(t), X'>h(t)\right\}}\\ = \sup_{z > h(t)} \frac{\bar F_V(z)}{\bar F_X(z)} {\mathbb{P}\left\{X+X'>t, X>h(t), X'>h(t)\right\}}\\ \end{multline}

where \(X,X' \sim F_X\). Multiplying by \(t\), we have

\begin{multline} t \mathbb{P}\left\{X\!+\!V>t, X\!>\!h(t), V\!>\!h(t) \right\} \le \sup_{z > h(t)} \frac{t\bar F_V(z)}{\bar F_X(z)} {\mathbb{P}\left\{X\!+\!X'>t, X\!>\!h(t), X'\!>\!h(t)\right\}}, \end{multline}

and because \(h(t) \to\infty\) as \(t \to\infty\) and \(\bar F_V(t) = O(\bar F_X(t) / t)\), we can say that for some \(c < \infty\), \(\lim_{t \to\infty} \sup_{z > h(t)} \frac{t\bar F_V(z)}{\bar F_X(z)} < c\). Therefore for sufficiently large t \(\mathbb{P}\left\{X+V>t, X>h(t), V>h(t) \right\} \leq \frac c t {\mathbb{P}\left\{X\!+\!X'>t, X\!>\!h(t), X'\!>\!h(t)\right\}}\). 

By Theorem 3.6, \(\mathbb{P}\left\{X\!+\!X'>t, X\!>\!h(t), X'\!>\!h(t)\right\}\) is \(o(\bar F_X(t))\), so the LHS is \(o(\bar F_X(t)/t)\) as desired.
\end{proof}

\subsubsection{Bounds on the numerator}

We want to show, for arbitrary \(\epsilon>0\), that \(\left|\int_{-\infty}^\infty vf_V(v)Q(v) - \int_{-\infty}^\infty vf_V(v)\right|<\epsilon\) in the limit as \(t\to\infty\). Since \(\left|\int_{-\infty}^\infty vf_V(v)Q(v) - \int_{-\infty}^\infty vf_V(v)\right| \le \int_{-\infty}^\infty \left|vf_V(v)(Q(v) - 1)\right| = \int_{-\infty}^\infty |v| \cdot f_V(v) \cdot |Q(v) - 1|\) it will suffice to show that the latter quantity is less than \(\epsilon\) for large \(t\).

We're going to show that \(\int_{-\infty}^\infty |v|\cdot f_V(v)\cdot |Q(v) - 1|\) is small by showing that the integral gets arbitrarily small on each of four pieces: \((-\infty,-h(t)]\), \((-h(t),h(t))\), \([h(t),t-h(t)]\), and \((t-h(t),\infty)\).

We'll handle these case by case (they'll get monotonically trickier).

\paragraph{Region 1: \texorpdfstring{\((-\infty,-h(t)]\)}{}}
 Since \(\int_{-\infty}^\infty vf_V(v)\) is absolutely convergent, for sufficiently large \(t\) we will have \(\int_{-\infty}^{-h(t)} |v|f_V(v) < \epsilon\), since \(h(t)\) goes to infinity by Lemma 1(a).

Since \(Q(v)\) is monotonically increasing and \(Q(0)=1\), we know that in this interval \(|Q(v)-1| = 1-Q(v)\).

So we have \(\int_{-\infty}^{-h(t)}|v|\cdot f_V(v) \cdot |Q(v)-1| = \int_{-\infty}^{-h(t)}|v|f_V(v)(1-Q(v)) < \int_{-\infty}^{-h(t)}|v|f_V(v) < \epsilon\) as desired.

\paragraph{Region 2: \texorpdfstring{\((-h(t),h(t))\)}{}}
 By lemma 1(d), \(h\) is such that for sufficiently large \(t\), \(|Q(v)-1|<\frac\epsilon{\int_{-\infty}^\infty |v|f_V(v)}\) on the interval \([-h(t),h(t)]\). (Note that the value of this upper bound depends only on \(V\) and \(\epsilon\), not on \(t\) or \(h\).) So we have \(\int_{-h(t)}^{h(t)}|v|f_V(v)|Q(v)-1| < \frac{\epsilon}{\int_{-\infty}^\infty|v|f_V(v)}\int_{-h(t)}^{h(t)}|v|f_V(v) < \frac{\epsilon}{\int_{-\infty}^\infty|v|f_V(v)}\int_{-\infty}^\infty|v|f_V(v) = \epsilon\).

\paragraph{Region 3: \texorpdfstring{\([h(t),t-h(t)]\)}{}}
 For the third part, we'd like to show that \(\int_{h(t)}^{t-h(t)}vf_V(v)(Q(v)-1)<\epsilon\). Since \(\int_{h(t)}^{t-h(t)}vf_V(v)(Q(v)-1) < \int_{h(t)}^{t-h(t)}tf_V(v)Q(v) = \frac t{\bar F_X(t)}\int_{h(t)}^{t-h(t)}f_V(v)\bar F_X(t-v)\) it would suffice to show that the latter expression becomes less than \(\epsilon\) for large \(t\), or equivalently that \(\int_{h(t)}^{t-h(t)}f_V(v)\bar F_X(t-v) = o\left(\frac{\bar F_X(t)}{t}\right)\).

The LHS in this expression is the unconditional probability that \(X+V>t\) and \(h(t)<V<t-h(t)\), but this event implies \(X+V>t, V>h(t)\), and \(X>h(t)\). So we can write

\begin{multline*}
\int_{h(t)}^{t-h(t)}f_V(v)\bar F_X(t-v) = Pr[X+V>t,\ h(t) < V < t - h(t)]
\\ < Pr[X+V>t,\ V > h(t),\ X > h(t)] = o(\bar F_X(t)/t)
\end{multline*}
by Lemma 2.

\paragraph{Region 4: \texorpdfstring{\((t-h(t),\infty)\)}{}}

For the fourth part, we'd like to show that \(\int_{t-h(t)}^\infty vf_V(v)Q(v) \to 0\) forlarge \(t\).

Since \(Q(v)=\frac{\bar F_X(t-v)}{\bar F_X(t)} < \frac 1{\bar F_X(t)}\), it would suffice to show \(\int_{t-h(t)}^\infty vf_V(v) = o(\bar F_X(t))\). But note that since \(\lim_{t\to\infty}\frac{\bar F_X(t-h(t))}{\bar F_X(t)} = 1\) by Lemma 1(c), this is equivalent to \(\int_{t-h(t)}^\infty vf_V(v) = o(\bar F_X(t-h(t)))\), which (by Lemma 1(b)) is equivalent to \(\int_t^\infty vf_V(v) = o(\bar F_X(t))\).

Note that \(\int_{t}^\infty vf_V(v) = t\int_{t}^\infty f_V(v) + \int_{t}^\infty (v-t)f_V(v) = t\bar F_V(t)+\int_{t}^\infty \bar F_V(v)\), so it will suffice to show that both terms in this sum are \(o(\bar F_X(t))\). 

The first term \(t \bar F_V(t)\) is \(o(\bar F_X(t))\) because we assumed \(\lim_{t\to\infty} \frac{t^p\bar F_V(t)}{\bar F_X(t)}=0\) for some \(p>1\).

For the second term, we have for the same reason \(\int_t^\infty \bar F_V(v) < \int_t^\infty \frac{\bar F_X(v)}{v^p} = \bar F_X(t)\int_t^\infty v^{-p} = \frac{t^{1-p}}{p-1}\bar F_X(t) = o(\bar F_X(t))\).

\subsubsection{Bounds on the denominator}

For the denominator, we want to show that \(\lim_{t\to\infty}\int_{-\infty}^\infty f_V(v)Q_t(v)=1=\int_{-\infty}^\infty f_V(v)\), so it'll suffice to show \(|\int_{-\infty}^\infty f_V(v)(Q_t(v)-1)|=o(1)\) as \(t\to\infty\). Again, we'll break up this integral into pieces, though they'll be more straightforward than last time. We'll look at \((-\infty,-h(t))\), \([-h(t),h(t)]\), and \((h(t),\infty)\).

\begin{itemize}
    \item \(|\int_{-\infty}^{-h(t)}f_V(v)(Q(v)-1)|=\int_{-\infty}^{-h(t)}f_V(v)(1-Q(v))<\int_{-\infty}^{-h(t)}f_V(v)\).
    \begin{itemize}
        \item But since \(h(t)\) goes to infinity, this left tail of the integral will contain less and less of \(V\) 's probability mass as $t$ increases.
    \end{itemize}
    \item \(|\int_{-h(t)}^{h(t)}f_V(v)(Q(v)-1)|\le\int_{-h(t)}^{h(t)}f_V(v)|Q(v)-1|\)
    \item \(\le \sup_{|v| \le h(t)} |Q(v,t)-1|\int_{-h(t)}^{h(t)}f_V(v)\le \sup_{|v| \le h(t)} |Q(v,t)-1|\)
    \begin{itemize}
        \item By Lemma 1(d) we know that this goes to zero for large \(t\).
    \end{itemize}
    \item \(|\int_{h(t)}^\infty f_V(v)(Q(v)-1)| = \int_{h(t)}^\infty f_V(v)(Q(v)-1) < \int_{h(t)}^\infty f_V(v)Q(v)\).
\end{itemize}

But for sufficiently large \(t\) we have \(h(t)>1\), so we obtain
 \(\int_{h(t)}^\infty f_V(v)Q(v)<\int_{h(t)}^\infty v f_V(v)Q(v) < \int_{-\infty}^\infty v f_V(v)Q(v) = o(1)\) 
by the results of the previous section. This completes the proof.

\subsection{Theorem 6}
\newtheorem*{theorem6}{Restatement of theorem \ref{thm6}}
\begin{theorem6}
    Let $X, V$ be independent random variables such that $\lim_{t\to\infty}\frac{\bar{F}_X(t+1)}{\bar{F}_X(t)}=0$. (This implies that X has tails that are dominated by $e^{-cx}$ for any c, though it's a slightly stronger claim because it requires that X not have large jumps in the decay of its tails.)
    Then for any V with a finite mean which has no upper bound, $\lim_{t\to\infty}\mathbb{E}[V|X+V > t] = \infty$. 
\end{theorem6}

Theorem \ref{thm6} generalizes a consequence of the "Regressional Goodhart Identity" in \citep{gao2023scaling}.

\begin{proof}
Let \(\text{Pr}(V>c+1)=p>0\), which exists by our assumption that \(V\) is unbounded.

Let \(\mathbb E[V|V<c] = q\). (If this is undefined because the conditional has probability \(0\), we'll have the desired result anyway since then \(V\) would always be at least \(c\).) 

Observe that for all \(t\), \(\mathbb E[V|V<c, X+V>t] \ge q\) (assuming it is defined), because we're conditioning \((V|V<c)\) on an event which is more likely for larger \(v\) (since \(X\) and \(V\) are independent). 

First, let's see that \(\lim_{t\to\infty}\frac{P(V<c|X+V\ge t)}{P(V>c+1|X+V\ge t)}=0\). This ratio of probabilities is equal to

\(\frac{\int_{-\infty}^c f_V(v)\bar F_X(t-v)}{\int_{c+1}^\infty f_V(v)\bar F_X(t-v)} \le \frac{\int_{-\infty}^c f_V(v)\bar F_X(t-c)}{\int_{c+1}^\infty f_V(v)\bar F_X(t-c-1)} = \frac{\bar F_X(t-c)}{\bar F_X(t-c-1)}\cdot \frac{\int_{-\infty}^c f_V(v)}{\int_{c+1}^\infty f_V(v)}\)

\(=\frac{\bar F_X(t-c)}{\bar F_X(t-c-1)}\cdot \frac{\text{Pr}(V<c)}{\text{Pr}(V>c+1)}\le \frac{\bar F_X(t-c)}{\bar F_X(t-c-1)}\cdot \frac1p\)

which, by our assumption that \(\lim_{t\to\infty}\frac{\bar{F}_X(t+1)}{\bar{F}_X(t)}=0\), will get arbitrarily small as \(t\) increases for any positive \(p\).

Now, consider \(\mathbb E[V|X+V\ge t]\). We can break this up as the sum across outcomes \(Z\) of \(\mathbb E[V|Z,X+V\ge t]\cdot \text{Pr}(Z | X+V\ge t)\) for the three disjoint outcomes \(V<c\), \(c\le V\le c+1\), and \(V>c+1\). Note that we can lower bound these expectations by \(q, c, c+1\) respectively. But then once \(t\) is large enough that \(\frac{\text{Pr}(V<c|X+V\ge t)}{\text{Pr}(V>c+1|X+V\ge t)}<\frac1{c-q}\),  this weighted sum of conditional expectations will add to more than \(c\).
\end{proof}

\setcounter{figure}{0}
\setcounter{table}{0}
\section{Additional experiments and figures}

\begin{figure}
    \centering
    \includegraphics[width=0.4\linewidth]{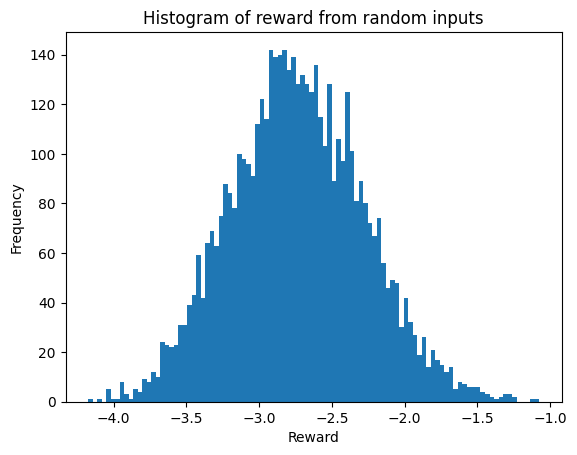}
    \includegraphics[width=0.4\linewidth]{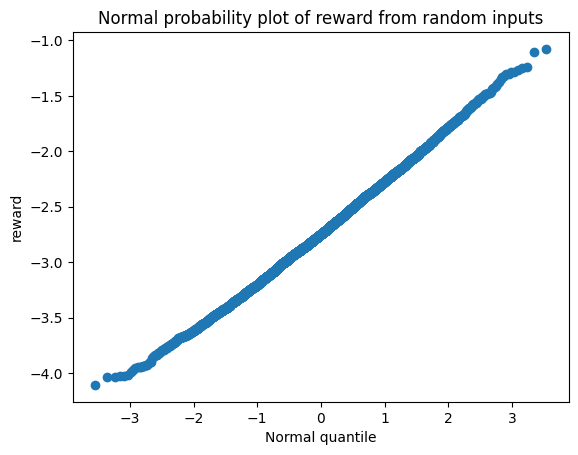}
    \caption{Histogram and normal probability plot of reward assigned by Pythia RM to random length-1024 token sequences. The Q-Q plot suggests the distribution is approximately normal, which is much lighter-tailed than exponential.}
    \label{fig:pythia-random}
\end{figure}

\begin{figure}
    \centering
    \includegraphics[width=0.5\linewidth]{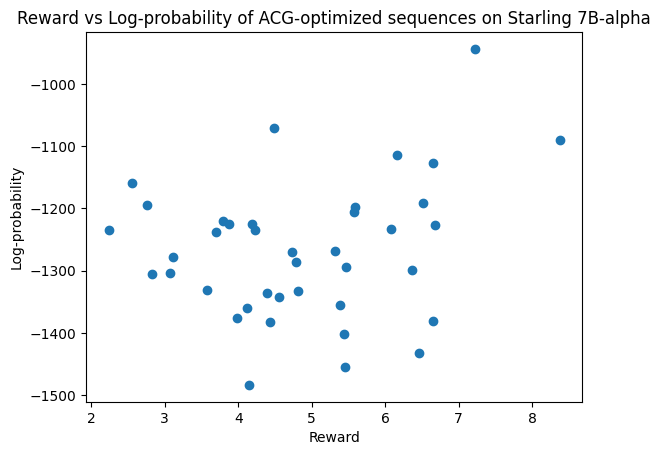}
    \caption{Reward and log-probability for ACG-optimized inputs to Starling 7B-alpha.}
    \label{fig:acg-results}
\end{figure}

Figures \ref{fig:pythia-random}, \ref{fig:acg-results} relate to experiments mentioned in the main paper.
In response to reviewer feedback, we added two further experiments to demonstrate the catastrophic Goodhart phenemonon with artificially heavy-tailed reward, one using best-of-N on synthetic distributions and one with PPO on Pythia 1B.

\subsection{Best-of-N experiment}

We created a synthetic experiment by letting reward $U = X + V$, where $X$ and $V$ are independent and sampled from different probability distributions, consistent with our theoretical assumptions. We vary $N$ from 1 to 65536, do 100 trials of taking the best-of-$N$ sample with highest $U$, and note whether $V$ goes towards 0 (overoptimization) or not.

Possible distributions for V are normal and t-distribution with df=10.
Possible distributions for X are normal, t with df=3, t with df=5, lognormal, and Levy. (All of these heavy-tailed except for the normal distribution.)
V is scaled to a standard deviation of 2 and X has s.d. of 1 (except for the Levy distribution, which has infinite variance), representing that in ordinary regimes most of the variance comes from utility rather than error.

\begin{figure}
    \centering
    \makebox[\textwidth][c]{\includegraphics[width=1.5\linewidth]{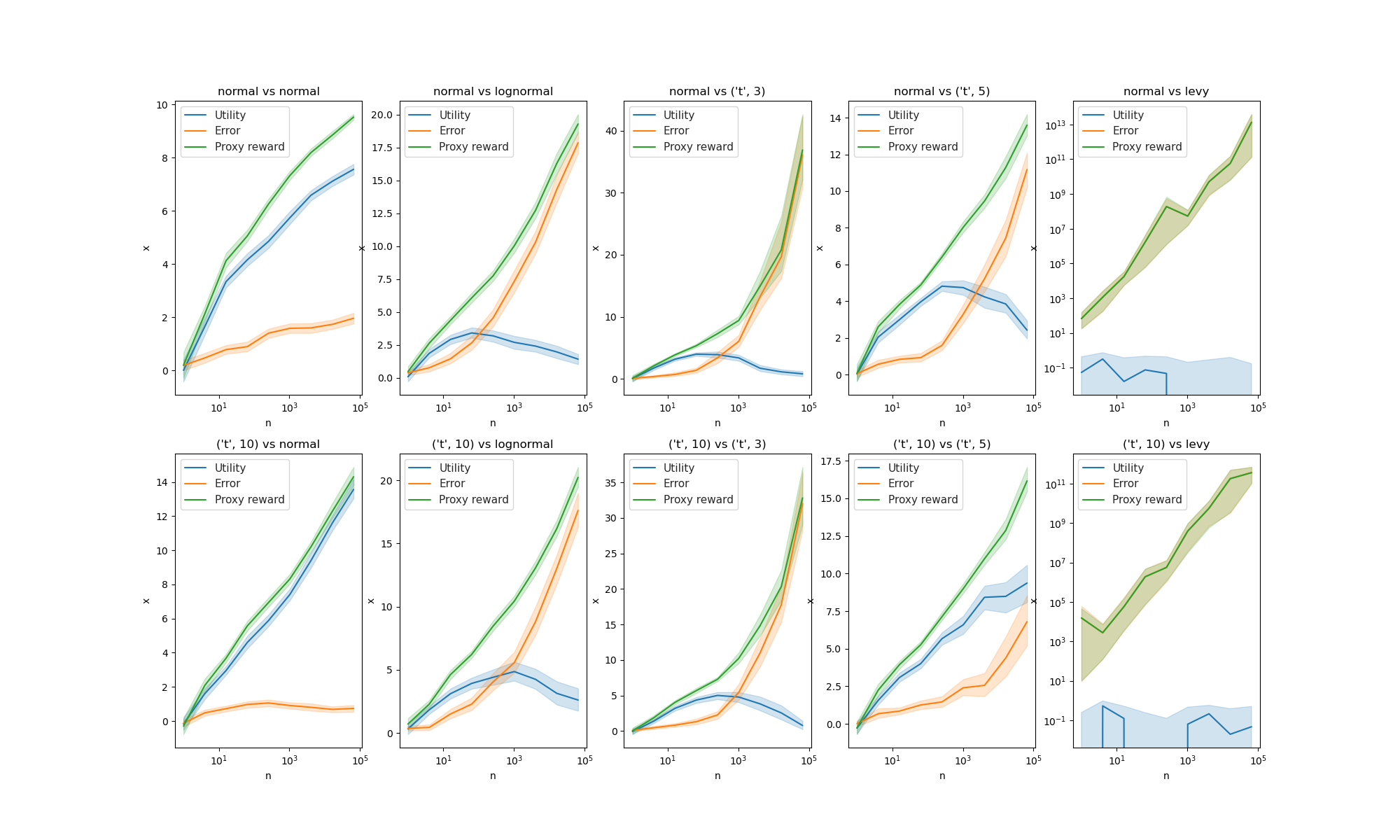}}
    \caption{When the error $X$ is normal and thus light-tailed, $V$ increases monotonically with $N$, consistent with our Theorem \ref{thm6}.
    However, when both $X$ and $V$ are heavy-tailed, we see results consistent with theorem \ref{thm5}. In $5$ of $6$ cases when $X$ is lognormal or student-t, $V$ first increases then starts to decline around $N=10^2$ or $10^3$. When $X$ is (t, df=5) and V is (t, df=10), $V$ instead peaks around $N=10^5$ (but declines afterwards). Finally, when X is Levy-distributed, utility never goes significantly above zero (optimization completely fails) because the Levy distribution is too heavy-tailed.}
    \label{fig:best-of-n}
\end{figure}

The results are shown in Figure \ref{fig:best-of-n}. Briefly, the results are consistent with the asymptotic results in theorems \ref{thm5} and \ref{thm6}, showing that overoptimization

\subsection{PPO experiment}

In this experiment, we examined PPO with artificially heavy-tailed rewards to see if catastrophic Goodhart could be observed.

OpenRLHF \citep{hu2024openrlhf} was used to train Pythia 1B with a reward model derived also from Pythia 1B, on the default OpenRLHF prompt dataset.
We used the reward model to represent true utility, and a heavy-tailed error term based on the number of "the" tokens was added to get the proxy reward.
The \texttt{kl\_target=0.5} option was used to dynamically adjust KL penalty, as we mention is done in \citet{ziegler2020finetuning}.
Rewards were not clipped. (Reward clipping can be useful to prevent overoptimization, but is not always used in PPO.)
Response length was limited to 256.

Two example completions are in figure \ref{fig:ppo}. Midway through training, the model generates reasonable text with a higher number of "the" tokens than usual, but as training progresses the model starts to generate completions with a very high number of "the" tokens while proxy reward reaches values $>10^9$, demonstrating that the basic pattern of catastrophic Goodhart can occur in RLHF under conditions of heavy-tailed error.

The result depends on hyperparameters (e.g. reward clipping would prevent this), so our observation should not be taken as a claim that catastrophic Goodhart is inevitable in all RLHF settings with heavy-tailed error.

\begin{figure}
    \centering
    \begin{framed}
        The preparation of polyarylethylene which is either employed for the purpose of applications in the petroleum refining or in the production of the well-known automobile oil lubes is carried out by the so-called preparation by the salt bath method described...
        in 1980 by D. W. Perkins in the United States. This method presupposes the usage of the salt bath as the medium for the preparation of the polyarylediene--the method of preparation of the crude oil base by the process which consists in letting it be saturated up with a particular the 
    \end{framed}
    \begin{framed}
        A mirror, the essence of the understanding of the given the conclusion that the manifestation in the form of the gender and the identification the solutions related to different fields of the modern utilization of the the functionality of the the the the modification of the the the the the the state of the the the the the the the the the the the employment and the the the the the the the the the the the the the the the the the the the the the the the the the the the the the the the the the the the the the and the the the the the the the the the the the the the the the the the the the the by the the the the the the the the the the the the the the the the the the the the the the the the the the the the the the the the the the the and the the the the the the the the the the the the the the the the the 
    \end{framed}
    \caption{PPO sample generations. Top: early in training, the model generates reasonable completions. Bottom: later in training, the artifically heavy-tailed reward dominates and the model generates completions with a very high number of "the" tokens.}
    \label{fig:ppo}
\end{figure}

\section{Assets}

We use three models for our experiments: Starling 7B-alpha, Llama 2 7B-chat, and Pythia-1.4B. Starling was developed by Berkeley, and Pythia by EleutherAI. Starling and Pythia models are licensed under Apache-2.0.\footnote{https://huggingface.co/berkeley-nest/Starling-RM-7B-alpha} \footnote{https://huggingface.co/EleutherAI/pythia-1.4b} Llama 2 models were developed by Meta and licensed under a license published by Meta.\footnote{https://ai.meta.com/llama/license/} 


\newpage
\section*{NeurIPS Paper Checklist}

The checklist is designed to encourage best practices for responsible machine learning research, addressing issues of reproducibility, transparency, research ethics, and societal impact. Do not remove the checklist: {\bf The papers not including the checklist will be desk rejected.} The checklist should follow the references and follow the (optional) supplemental material.  The checklist does NOT count towards the page
limit. 

Please read the checklist guidelines carefully for information on how to answer these questions. For each question in the checklist:
\begin{itemize}
    \item You should answer \answerYes{}, \answerNo{}, or \answerNA{}.
    \item \answerNA{} means either that the question is Not Applicable for that particular paper or the relevant information is Not Available.
    \item Please provide a short (1–2 sentence) justification right after your answer (even for NA). 
\end{itemize}

{\bf The checklist answers are an integral part of your paper submission.} They are visible to the reviewers, area chairs, senior area chairs, and ethics reviewers. You will be asked to also include it (after eventual revisions) with the final version of your paper, and its final version will be published with the paper.

The reviewers of your paper will be asked to use the checklist as one of the factors in their evaluation. While "\answerYes{}" is generally preferable to "\answerNo{}", it is perfectly acceptable to answer "\answerNo{}" provided a proper justification is given (e.g., "error bars are not reported because it would be too computationally expensive" or "we were unable to find the license for the dataset we used"). In general, answering "\answerNo{}" or "\answerNA{}" is not grounds for rejection. While the questions are phrased in a binary way, we acknowledge that the true answer is often more nuanced, so please just use your best judgment and write a justification to elaborate. All supporting evidence can appear either in the main paper or the supplemental material, provided in appendix. If you answer \answerYes{} to a question, in the justification please point to the section(s) where related material for the question can be found.

IMPORTANT, please:
\begin{itemize}
    \item {\bf Delete this instruction block, but keep the section heading ``NeurIPS paper checklist"},
    \item  {\bf Keep the checklist subsection headings, questions/answers and guidelines below.}
    \item {\bf Do not modify the questions and only use the provided macros for your answers}.
\end{itemize}


\begin{enumerate}

\item {\bf Claims}
    \item[] Question: Do the main claims made in the abstract and introduction accurately reflect the paper's contributions and scope?
    \item[] Answer: \answerYes{} 
    \item[] Justification: The abstract lists the important claims: the relationship between Goodhart's Law and whether the error in a misspecified reward is heavy-tailed. The main limitation of independence assumptions is clearly stated in the introduction.
    \item[] Guidelines:
    \begin{itemize}
        \item The answer NA means that the abstract and introduction do not include the claims made in the paper.
        \item The abstract and/or introduction should clearly state the claims made, including the contributions made in the paper and important assumptions and limitations. A No or NA answer to this question will not be perceived well by the reviewers. 
        \item The claims made should match theoretical and experimental results, and reflect how much the results can be expected to generalize to other settings. 
        \item It is fine to include aspirational goals as motivation as long as it is clear that these goals are not attained by the paper. 
    \end{itemize}

\item {\bf Limitations}
    \item[] Question: Does the paper discuss the limitations of the work performed by the authors?
    \item[] Answer: \answerYes{} 
    \item[] Justification: Section 7 lists the limitations, which we have combined with the discussion section due to heavy overlap.
    \item[] Guidelines:
    \begin{itemize}
        \item The answer NA means that the paper has no limitation while the answer No means that the paper has limitations, but those are not discussed in the paper. 
        \item The authors are encouraged to create a separate "Limitations" section in their paper.
        \item The paper should point out any strong assumptions and how robust the results are to violations of these assumptions (e.g., independence assumptions, noiseless settings, model well-specification, asymptotic approximations only holding locally). The authors should reflect on how these assumptions might be violated in practice and what the implications would be.
        \item The authors should reflect on the scope of the claims made, e.g., if the approach was only tested on a few datasets or with a few runs. In general, empirical results often depend on implicit assumptions, which should be articulated.
        \item The authors should reflect on the factors that influence the performance of the approach. For example, a facial recognition algorithm may perform poorly when image resolution is low or images are taken in low lighting. Or a speech-to-text system might not be used reliably to provide closed captions for online lectures because it fails to handle technical jargon.
        \item The authors should discuss the computational efficiency of the proposed algorithms and how they scale with dataset size.
        \item If applicable, the authors should discuss possible limitations of their approach to address problems of privacy and fairness.
        \item While the authors might fear that complete honesty about limitations might be used by reviewers as grounds for rejection, a worse outcome might be that reviewers discover limitations that aren't acknowledged in the paper. The authors should use their best judgment and recognize that individual actions in favor of transparency play an important role in developing norms that preserve the integrity of the community. Reviewers will be specifically instructed to not penalize honesty concerning limitations.
    \end{itemize}

\item {\bf Theory Assumptions and Proofs}
    \item[] Question: For each theoretical result, does the paper provide the full set of assumptions and a complete (and correct) proof?
    \item[] Answer: \answerYes{} 
    \item[] Justification: All proofs are given inline or in the appendix, except for Theorem 5 which appears in the supplemental material.
    \item[] Guidelines:
    \begin{itemize}
        \item The answer NA means that the paper does not include theoretical results. 
        \item All the theorems, formulas, and proofs in the paper should be numbered and cross-referenced.
        \item All assumptions should be clearly stated or referenced in the statement of any theorems.
        \item The proofs can either appear in the main paper or the supplemental material, but if they appear in the supplemental material, the authors are encouraged to provide a short proof sketch to provide intuition. 
        \item Inversely, any informal proof provided in the core of the paper should be complemented by formal proofs provided in appendix or supplemental material.
        \item Theorems and Lemmas that the proof relies upon should be properly referenced. 
    \end{itemize}

    \item {\bf Experimental Result Reproducibility}
    \item[] Question: Does the paper fully disclose all the information needed to reproduce the main experimental results of the paper to the extent that it affects the main claims and/or conclusions of the paper (regardless of whether the code and data are provided or not)?
    \item[] Answer: \answerYes{} 
    \item[] Justification: Sampling rewards requires no hyperparameters, and hyperparameters are provided for ACG.
    \item[] Guidelines:
    \begin{itemize}
        \item The answer NA means that the paper does not include experiments.
        \item If the paper includes experiments, a No answer to this question will not be perceived well by the reviewers: Making the paper reproducible is important, regardless of whether the code and data are provided or not.
        \item If the contribution is a dataset and/or model, the authors should describe the steps taken to make their results reproducible or verifiable. 
        \item Depending on the contribution, reproducibility can be accomplished in various ways. For example, if the contribution is a novel architecture, describing the architecture fully might suffice, or if the contribution is a specific model and empirical evaluation, it may be necessary to either make it possible for others to replicate the model with the same dataset, or provide access to the model. In general. releasing code and data is often one good way to accomplish this, but reproducibility can also be provided via detailed instructions for how to replicate the results, access to a hosted model (e.g., in the case of a large language model), releasing of a model checkpoint, or other means that are appropriate to the research performed.
        \item While NeurIPS does not require releasing code, the conference does require all submissions to provide some reasonable avenue for reproducibility, which may depend on the nature of the contribution. For example
        \begin{enumerate}
            \item If the contribution is primarily a new algorithm, the paper should make it clear how to reproduce that algorithm.
            \item If the contribution is primarily a new model architecture, the paper should describe the architecture clearly and fully.
            \item If the contribution is a new model (e.g., a large language model), then there should either be a way to access this model for reproducing the results or a way to reproduce the model (e.g., with an open-source dataset or instructions for how to construct the dataset).
            \item We recognize that reproducibility may be tricky in some cases, in which case authors are welcome to describe the particular way they provide for reproducibility. In the case of closed-source models, it may be that access to the model is limited in some way (e.g., to registered users), but it should be possible for other researchers to have some path to reproducing or verifying the results.
        \end{enumerate}
    \end{itemize}

\item {\bf Open access to data and code}
    \item[] Question: Does the paper provide open access to the data and code, with sufficient instructions to faithfully reproduce the main experimental results, as described in supplemental material?
    \item[] Answer: \answerYes{} 
    \item[] Justification: Code will be provided in the supplemental material.
    \item[] Guidelines:
    \begin{itemize}
        \item The answer NA means that paper does not include experiments requiring code.
        \item Please see the NeurIPS code and data submission guidelines (\url{https://nips.cc/public/guides/CodeSubmissionPolicy}) for more details.
        \item While we encourage the release of code and data, we understand that this might not be possible, so “No” is an acceptable answer. Papers cannot be rejected simply for not including code, unless this is central to the contribution (e.g., for a new open-source benchmark).
        \item The instructions should contain the exact command and environment needed to run to reproduce the results. See the NeurIPS code and data submission guidelines (\url{https://nips.cc/public/guides/CodeSubmissionPolicy}) for more details.
        \item The authors should provide instructions on data access and preparation, including how to access the raw data, preprocessed data, intermediate data, and generated data, etc.
        \item The authors should provide scripts to reproduce all experimental results for the new proposed method and baselines. If only a subset of experiments are reproducible, they should state which ones are omitted from the script and why.
        \item At submission time, to preserve anonymity, the authors should release anonymized versions (if applicable).
        \item Providing as much information as possible in supplemental material (appended to the paper) is recommended, but including URLs to data and code is permitted.
    \end{itemize}

\item {\bf Experimental Setting/Details}
    \item[] Question: Does the paper specify all the training and test details (e.g., data splits, hyperparameters, how they were chosen, type of optimizer, etc.) necessary to understand the results?
    \item[] Answer: \answerYes{} 
    \item[] Justification: Other than hyperparameters, there are no details required to understand the results.
    \item[] Guidelines:
    \begin{itemize}
        \item The answer NA means that the paper does not include experiments.
        \item The experimental setting should be presented in the core of the paper to a level of detail that is necessary to appreciate the results and make sense of them.
        \item The full details can be provided either with the code, in appendix, or as supplemental material.
    \end{itemize}

\item {\bf Experiment Statistical Significance}
    \item[] Question: Does the paper report error bars suitably and correctly defined or other appropriate information about the statistical significance of the experiments?
    \item[] Answer: \answerYes{} 
    \item[] Justification: The only error bars are inter-run variability of ACG. The standard deviation was reported rather than error bars due to the small number of runs.
    \item[] Guidelines:
    \begin{itemize}
        \item The answer NA means that the paper does not include experiments.
        \item The authors should answer "Yes" if the results are accompanied by error bars, confidence intervals, or statistical significance tests, at least for the experiments that support the main claims of the paper.
        \item The factors of variability that the error bars are capturing should be clearly stated (for example, train/test split, initialization, random drawing of some parameter, or overall run with given experimental conditions).
        \item The method for calculating the error bars should be explained (closed form formula, call to a library function, bootstrap, etc.)
        \item The assumptions made should be given (e.g., Normally distributed errors).
        \item It should be clear whether the error bar is the standard deviation or the standard error of the mean.
        \item It is OK to report 1-sigma error bars, but one should state it. The authors should preferably report a 2-sigma error bar than state that they have a 96\% CI, if the hypothesis of Normality of errors is not verified.
        \item For asymmetric distributions, the authors should be careful not to show in tables or figures symmetric error bars that would yield results that are out of range (e.g. negative error rates).
        \item If error bars are reported in tables or plots, The authors should explain in the text how they were calculated and reference the corresponding figures or tables in the text.
    \end{itemize}

\item {\bf Experiments Compute Resources}
    \item[] Question: For each experiment, does the paper provide sufficient information on the computer resources (type of compute workers, memory, time of execution) needed to reproduce the experiments?
    \item[] Answer: \answerYes{} 
    \item[] Justification: The experiments took minimal compute resources except H100 hours for ACG, and we report the number of GPU-hours used.
    \item[] Guidelines:
    \begin{itemize}
        \item The answer NA means that the paper does not include experiments.
        \item The paper should indicate the type of compute workers CPU or GPU, internal cluster, or cloud provider, including relevant memory and storage.
        \item The paper should provide the amount of compute required for each of the individual experimental runs as well as estimate the total compute. 
        \item The paper should disclose whether the full research project required more compute than the experiments reported in the paper (e.g., preliminary or failed experiments that didn't make it into the paper). 
    \end{itemize}
    
\item {\bf Code Of Ethics}
    \item[] Question: Does the research conducted in the paper conform, in every respect, with the NeurIPS Code of Ethics \url{https://neurips.cc/public/EthicsGuidelines}?
    \item[] Answer: \answerYes{} 
    \item[] Justification: The research conforms to all data-related concerns. No human subjects were involved, and we think the risk of harmful societal impact is minimal.
    \item[] Guidelines:
    \begin{itemize}
        \item The answer NA means that the authors have not reviewed the NeurIPS Code of Ethics.
        \item If the authors answer No, they should explain the special circumstances that require a deviation from the Code of Ethics.
        \item The authors should make sure to preserve anonymity (e.g., if there is a special consideration due to laws or regulations in their jurisdiction).
    \end{itemize}

\item {\bf Broader Impacts}
    \item[] Question: Does the paper discuss both potential positive societal impacts and negative societal impacts of the work performed?
    \item[] Answer: \answerYes{} 
    \item[] Justification: The immediate societal impacts are limited, but we discuss some potential applications to long-term safety.
    \item[] Guidelines:
    \begin{itemize}
        \item The answer NA means that there is no societal impact of the work performed.
        \item If the authors answer NA or No, they should explain why their work has no societal impact or why the paper does not address societal impact.
        \item Examples of negative societal impacts include potential malicious or unintended uses (e.g., disinformation, generating fake profiles, surveillance), fairness considerations (e.g., deployment of technologies that could make decisions that unfairly impact specific groups), privacy considerations, and security considerations.
        \item The conference expects that many papers will be foundational research and not tied to particular applications, let alone deployments. However, if there is a direct path to any negative applications, the authors should point it out. For example, it is legitimate to point out that an improvement in the quality of generative models could be used to generate deepfakes for disinformation. On the other hand, it is not needed to point out that a generic algorithm for optimizing neural networks could enable people to train models that generate Deepfakes faster.
        \item The authors should consider possible harms that could arise when the technology is being used as intended and functioning correctly, harms that could arise when the technology is being used as intended but gives incorrect results, and harms following from (intentional or unintentional) misuse of the technology.
        \item If there are negative societal impacts, the authors could also discuss possible mitigation strategies (e.g., gated release of models, providing defenses in addition to attacks, mechanisms for monitoring misuse, mechanisms to monitor how a system learns from feedback over time, improving the efficiency and accessibility of ML).
    \end{itemize}
    
\item {\bf Safeguards}
    \item[] Question: Does the paper describe safeguards that have been put in place for responsible release of data or models that have a high risk for misuse (e.g., pretrained language models, image generators, or scraped datasets)?
    \item[] Answer: \answerNA{} 
    \item[] Justification: We have created no such artifacts.
    \item[] Guidelines:
    \begin{itemize}
        \item The answer NA means that the paper poses no such risks.
        \item Released models that have a high risk for misuse or dual-use should be released with necessary safeguards to allow for controlled use of the model, for example by requiring that users adhere to usage guidelines or restrictions to access the model or implementing safety filters. 
        \item Datasets that have been scraped from the Internet could pose safety risks. The authors should describe how they avoided releasing unsafe images.
        \item We recognize that providing effective safeguards is challenging, and many papers do not require this, but we encourage authors to take this into account and make a best faith effort.
    \end{itemize}

\item {\bf Licenses for existing assets}
    \item[] Question: Are the creators or original owners of assets (e.g., code, data, models), used in the paper, properly credited and are the license and terms of use explicitly mentioned and properly respected?
    \item[] Answer: \answerYes{} 
    \item[] Justification: The assets we use are Starling 7B-alpha, Llama 2 7B-chat, and Pythia-1.4B. Starling and Pythia models are licensed under Apache-2.0.\footnote{https://twitter.com/NexusflowX/status/1770532630645420474} \footnote{https://huggingface.co/EleutherAI/pythia-1.4b} Llama 2 models are licensed under a license published by Meta.\footnote{https://ai.meta.com/llama/license/} We are in compliance with all licenses and terms of use.
    \item[] Guidelines:
    \begin{itemize}
        \item The answer NA means that the paper does not use existing assets.
        \item The authors should cite the original paper that produced the code package or dataset.
        \item The authors should state which version of the asset is used and, if possible, include a URL.
        \item The name of the license (e.g., CC-BY 4.0) should be included for each asset.
        \item For scraped data from a particular source (e.g., website), the copyright and terms of service of that source should be provided.
        \item If assets are released, the license, copyright information, and terms of use in the package should be provided. For popular datasets, \url{paperswithcode.com/datasets} has curated licenses for some datasets. Their licensing guide can help determine the license of a dataset.
        \item For existing datasets that are re-packaged, both the original license and the license of the derived asset (if it has changed) should be provided.
        \item If this information is not available online, the authors are encouraged to reach out to the asset's creators.
    \end{itemize}

\item {\bf New Assets}
    \item[] Question: Are new assets introduced in the paper well documented and is the documentation provided alongside the assets?
    \item[] Answer: \answerNA{} 
    \item[] Justification: We do not release new assets.
    \item[] Guidelines:
    \begin{itemize}
        \item The answer NA means that the paper does not release new assets.
        \item Researchers should communicate the details of the dataset/code/model as part of their submissions via structured templates. This includes details about training, license, limitations, etc. 
        \item The paper should discuss whether and how consent was obtained from people whose asset is used.
        \item At submission time, remember to anonymize your assets (if applicable). You can either create an anonymized URL or include an anonymized zip file.
    \end{itemize}

\item {\bf Crowdsourcing and Research with Human Subjects}
    \item[] Question: For crowdsourcing experiments and research with human subjects, does the paper include the full text of instructions given to participants and screenshots, if applicable, as well as details about compensation (if any)? 
    \item[] Answer: \answerNA{} 
    \item[] Justification: No human subjects are involved.
    \item[] Guidelines:
    \begin{itemize}
        \item The answer NA means that the paper does not involve crowdsourcing nor research with human subjects.
        \item Including this information in the supplemental material is fine, but if the main contribution of the paper involves human subjects, then as much detail as possible should be included in the main paper. 
        \item According to the NeurIPS Code of Ethics, workers involved in data collection, curation, or other labor should be paid at least the minimum wage in the country of the data collector. 
    \end{itemize}

\item {\bf Institutional Review Board (IRB) Approvals or Equivalent for Research with Human Subjects}
    \item[] Question: Does the paper describe potential risks incurred by study participants, whether such risks were disclosed to the subjects, and whether Institutional Review Board (IRB) approvals (or an equivalent approval/review based on the requirements of your country or institution) were obtained?
    \item[] Answer: \answerNA{} 
    \item[] Justification: No human subjects are involved.
    \item[] Guidelines:
    \begin{itemize}
        \item The answer NA means that the paper does not involve crowdsourcing nor research with human subjects.
        \item Depending on the country in which research is conducted, IRB approval (or equivalent) may be required for any human subjects research. If you obtained IRB approval, you should clearly state this in the paper. 
        \item We recognize that the procedures for this may vary significantly between institutions and locations, and we expect authors to adhere to the NeurIPS Code of Ethics and the guidelines for their institution. 
        \item For initial submissions, do not include any information that would break anonymity (if applicable), such as the institution conducting the review.
    \end{itemize}

\end{enumerate}

\end{document}